\documentclass[journal]{IEEEtran}
\usepackage{lineno,hyperref}
\usepackage{bbm}
\usepackage{bm}
\usepackage{amsfonts}
\usepackage{amsmath}
\usepackage{mathrsfs}
\usepackage{amssymb}
\usepackage{enumerate}
\usepackage{graphicx}
\usepackage{subfigure}
\usepackage{booktabs}
\usepackage{graphicx}
\usepackage{setspace}
\usepackage{algorithm}
\usepackage{algorithmic}
\usepackage{setspace}
\usepackage{multirow}
\usepackage{color}
\usepackage{cite}
\usepackage{array}
\usepackage{times}
\usepackage{mathptmx}
\usepackage{dsfont}
\ifCLASSINFOpdf
\else
\fi
\begin{document}
\title{Progressive Feature Mining and External Knowledge-Assisted Text-Pedestrian Image Retrieval}
\author{Huafeng~Li, Shedan~Yang, Yafei Zhang, Dapeng Tao \IEEEmembership{}, Zhengtao~Yu 
\thanks{This work was supported in part by the National Natural Science Foundation of China under Grant 61966021, Grant 62276120, and the Yunnan Fundamental Research Projects (202301AV070004).}
\thanks{H. Li, S. Yang, Y. Zhang and Z. Yu are with the Faculty of Information Engineering and Automation, Kunming University of Science and Technology, Kunming 650500, China.(E-mail:lhfchina99@kust.edu.cn (H. Li); ysd199@163.com(S. Yang))}
\thanks{D. Tao is with FIST LAB, School of Information Science and Engineering, Yunnan University, Kunming 650091, China.}
\thanks{}
\thanks{ ${\ast}$ indicates contributed to this work equally.}
\thanks{Manuscript received xxxx;}}
\markboth{Journal of \LaTeX\ Class Files}%
{Shell \MakeLowercase{\textit{et al.}}}
\maketitle
\begin{abstract}
Text-Pedestrian Image Retrieval aims to use the text describing pedestrian appearance to retrieve the corresponding pedestrian image. This task involves not only modality discrepancy, but also the challenge of the textual diversity of pedestrians with the same identity. At present, although existing research progress has been made in text-pedestrian image retrieval, these methods do not comprehensively consider the above-mentioned problems. Considering these, this paper proposes a progressive feature mining and external knowledge-assisted feature purification method. Specifically, we use a progressive mining mode to enable the model to mine discriminative features from neglected information, thereby avoiding the loss of discriminative information and improving the expression ability of features. In addition, to further reduce the negative impact of modal discrepancy and text diversity on cross-modal matching, we propose to use other sample knowledge of the same modality, \emph{i.e.}, external knowledge to enhance identity-consistent features and weaken identity-inconsistent features. This process purifies features and alleviates the interference caused by textual diversity and negative sample correlation features of the same modal. Extensive experiments on three challenging datasets demonstrate the effectiveness and superiority of the proposed method, and the retrieval performance even surpasses that of the large-scale model-based method on large-scale datasets.
\end{abstract}
\IEEEpeerreviewmaketitle
\section{Introduction}
Person re-identification plays a vital role in social public safety, many scholars have conducted in-depth research on this task \cite{56,57,63,58,65,59,64,60}, and text-pedestrian image retrieval is one of the sub-tasks. Text-pedestrian image retrieval aims at retrieving pedestrian images in the surveillance system based on textual descriptions of pedestrian appearance, so as to achieve cross-camera tracking of target pedestrians. Currently, although surveillance cameras are widely set in various public places, there are still some areas that cannot be involved due to their particularity, which has caused the existence of monitoring blind areas. In this case, we can not retrieve the target pedestrian captured by the surveillance system based on the query image. However, if witnesses exist, the textual descriptions of target pedestrians can be obtained. Therefore, text-based pedestrian image retrieval is particularly significant.
\begin{figure}[t!]
\centering
\includegraphics{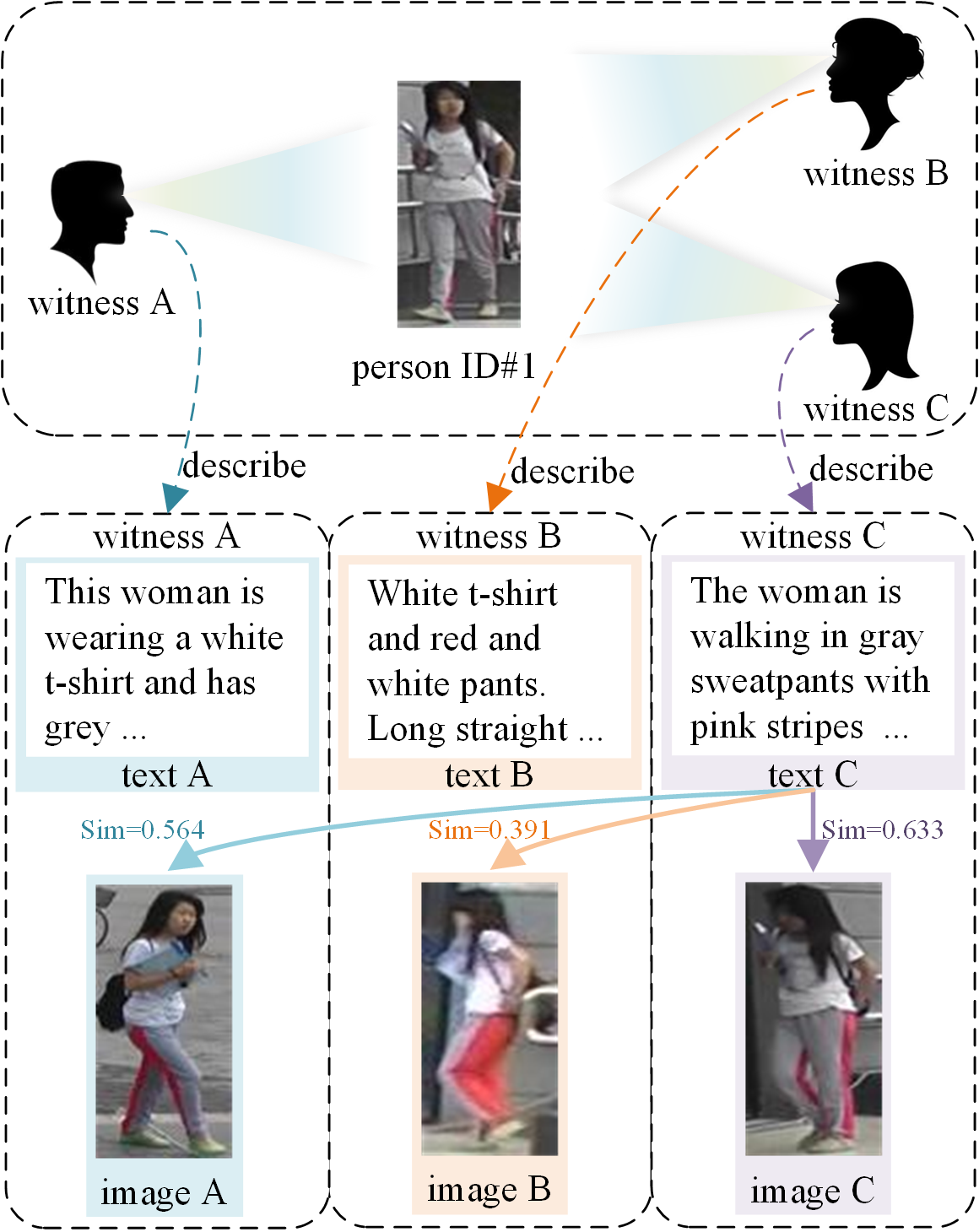}
\caption{Witnesses A, B, and C observed the target pedestrian from different perspectives and provided respective descriptions, which are three textual descriptions: text A, text B and text C. The similarities between these three texts and the image of the target pedestrian captured by the camera exhibit variations.}
\label{label1}
\end{figure}

In text-pedestrian image retrieval, the main challenge is to address the modality discrepancy problem. To tackle this issue, a series of effective methods have been proposed. These methods can be roughly divided into three types according to their characteristics: word-level relationship construction (WRC)-based methods \cite{1(7),2(8),3(10),4(12),5(18)}, cross-modal semantic alignment (CSA)-based methods \cite{6(4),7(11),8(37),9(42),10(47),11(57)}, and external model assistance (EMA)-based methods \cite{12(14),13(19),14(29),15(27),16(32)}. The WRC-based methods construct cross-modal correspondences by measuring the similarity between word-level features of text and region-level features of the image. The CSA-based methods align keywords or phrases from the text with corresponding regions in the image to endow text and image features with consistent semantic information, which helps to reduce misalignment caused by feature inconsistency. The EMA-based methods utilize auxiliary external models to extract specific text and image features. Common external models include keypoint estimation models \cite{12(14)}, pose estimation models \cite{14(29)}, human parsing models \cite{13(19)}, \emph{etc}. However, these methods require external models to acquire external knowledge, resulting in high reliance on external models. In addition, such methods do not take into account the impact of textual diversity of the same pedestrians on cross-modal retrieval.

As shown in Fig. \ref{label1}, the text descriptions of the same pedestrian exhibit significant differences due to the uniqueness of language expressions from different witnesses and various observation perspectives. If the model lacks the robustness to handle such diversity, it will limit the improvement of retrieval performance. To address the issue, an adversarial attack and defense method is proposed for text-based pedestrian image retrieval \cite{9(42)}. However, it overlooks the role of deep-level discriminative features, causing suboptimal retrieval performance, deep-level discriminative features are essential for recognition tasks \cite{55}. To tackle the aforementioned challenges, we develop a \textbf{p}rogressive \textbf{f}eature \textbf{m}ining and \textbf{e}xternal \textbf{k}nowledge-assisted \textbf{f}eature \textbf{p}urification (PFM-EKFP) method.
To make features from both text and image more expressive, we propose to further extract features from the information ignored by cross-modal identity consistency features, effectively preventing the loss of discriminative features and enhancing feature expression capability.  When dealing with sample diversity, we utilize the information from other samples with the same identity to assist in purifying the current sample's features, thereby enhancing identity-consistent shared features and weakening identity-irrelevant features. In this design, we leverage the fact that the text features associated with the same identity are invariant, while the identity-irrelevant features such as sentence style are variable.

Methodologically, the developed progressive feature mining (PFM) mechanism first extracts cross-modal consistent discriminative information from the features obtained by the backbone. Subsequently, under the guidance of consistency information, we design a dual reverse attention module (DRAM) and a single reverse attention module (SRAM). The DRAM is used to search for shared features between text and image of the same identity, as well as features other than shared features. We further feed the features other than shared features into the feature extraction network to extract discriminative features, and then combine them with the shared features output by the DRAM. The combined shared features are used to promote the original features to be more discriminative even after the combined shared features are suppressed. This PFM mechanism effectively enhances the model's ability to extract discriminative features, making the extracted features more expressive. In external knowledge-assisted feature representation (EKFR), we utilize text descriptions of the same pedestrian to highlight shared features and suppress inconsistent features. For pedestrian images, the proposed method no longer requires external knowledge from the same identity in order to achieve correlation suppression for similar features of different identities. Throughout this process, EKFR employs the same parameters for processing both text and image to facilitate the learning of modality-invariant features, and the image is not processed in the testing phase in EKFR.

In summary,  this paper has three main contributions as follows.
\begin{itemize}
\item A progressive discriminative feature mining mechanism is proposed, which compensates for the lack of shared features information by further extracting discriminative features from non-shared features, thereby improving the expression ability of features.

\item We devise an external knowledge-assisted feature purification strategy, which introduces other samples with the same identity to enhance identity-consistent features and weaken identity-inconsistent features, thereby purifying features and effectively solving the negative impact of text diversity on text-pedestrian image retrieval.

\item To validate the effectiveness of the proposed method, this study conducted experiments on three challenging datasets. The experimental results demonstrate that on large datasets, the performance of the proposed method not only reaches but even surpasses the performance achieved by methods based on vision-language pretraining models. This confirms the effectiveness of the proposed approach and its superiority over other methods.
\end{itemize}

The remaining part of this paper is organized as follows: Section \uppercase\expandafter{\romannumeral2} reviews some related work; Section \uppercase\expandafter{\romannumeral3} elaborates on the developed method; Section \uppercase\expandafter{\romannumeral4} validates the effectiveness of the method through experiments, and Section \uppercase\expandafter{\romannumeral5} summarizes this paper.

\begin{figure*}[t!]
\centering
\includegraphics[width=7.2in,height=3.4in]{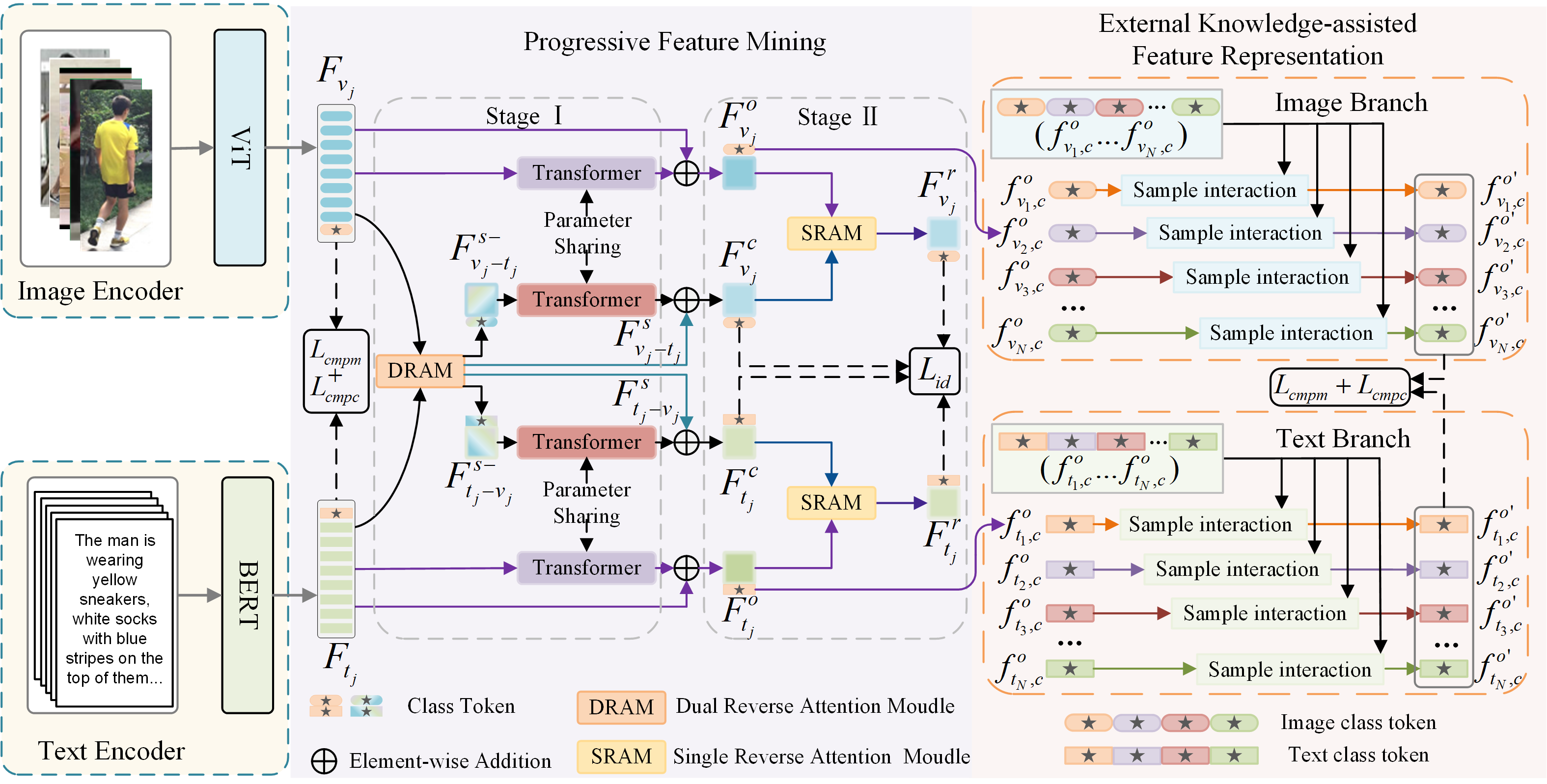}
\caption{Overall architecture of the proposed model. The input pedestrian image $\bm x_{v_{j}}$ and text $\bm x_{t_{j}}$ are extracted through ViT and BERT respectively to obtain features $\bm F_{v_{j}}$ and $\bm F_{t_{j}}$, respectively. Dual attention computation is performed on $\bm F_{v_{j}}$ and $\bm F_{t_{j}}$ to identify inconsistent features between them. Discriminative features from the remaining information are further extracted from the inconsistent features of $\bm F_{v_{j}}$ and $\bm F_{t_{j}}$ using Transformer. To achieve this purpose, reverse cross-attention computation is performed between the features extracted from the remaining information and the features extracted from $\bm F_{v_{j}}$ and $\bm F_{t_{j}}$, suppressing similar information and highlighting complementary information. Finally, $\bm F_{v_{j}}$ and $\bm F_{t_{j}}$ are combined with the final complementary information as the input of the external knowledge-assisted module. The external knowledge-assisted module leverages the features of other samples to assist the representation of current sample features, further mitigating modality disparities and enhancing the discriminative representation of the features.}
\label{label2}
\end{figure*}

\section{Related Work}
Since text-pedestrian image retrieval can be regarded as a sub-problem in text-image matching, in the introduction of related work, we first review the relevant work on text-image matching.
\subsection{Text-Image Matching}
Text-image matching aims to retrieve images that are consistent with the text description. In this task, the modal discrepancy between image and text is a critical factor affecting matching performance. To address this issue, Zheng $et \ al$. \cite{18(2-1)} proposed a dual-path CNN model for visual-textual embedding, extracting image and text features separately. They enforced the network with an instance loss function to generate feature representations suitable for image-text matching. Messina $et \ al$. \cite{19(2-2)} introduced a transformer encoder reasoning network capable of reasoning on features from different modalities independently and reducing the modal discrepancy by sharing the weights of the deeper transformer layers. However, the above-mentioned methods primarily rely on loss functions to narrow the gap between features from different modalities, and it is difficult to solve the challenges brought by the modal discrepancy. To address this, Ji $et \ al$. \cite{20(2-8),21(2-10)} proposed executing bidirectional interactions between image and text features to overcome modal discrepancy. Nevertheless, such method needs to consider all possible text-image pairs formed by the text and all the images, which increases the computational burden and hinders the deployment of the model in real-world scenarios.

In addition, the textual descriptions often contain multiple objects, and there may exist spatial relationships among these objects. Therefore, how to leverage the spatial position information of the objects to achieve text-image matching is considered a key factor in improving matching performance. To address this, Li $et \ al$. \cite{22(2-14)} proposed constructing connections between image regions and performed reasoning using Graph Convolutional Networks to generate visual features that have consistent semantic relationships with text descriptions, thereby achieving cross-modal matching. Liu $et \ al$. \cite{23(2-13)} introduced a relationship-enhanced semantic graph model, which learns concept representations of image local regions and text words, and then constructs image regions relationships and text words relationships in a contextual order to optimize image and text representations. Ge $et \ al$. \cite{24(2-16)} proposed a cross-modal semantic enhanced interaction method, which improves feature representation accuracy by establishing cross-modal semantic associations between objects and corresponding words. Although the above methods are effective, they may face challenges when directly applied to text-pedestrian image retrieval and may not achieve superior performance. This is because the textual description of pedestrians is often only aimed at a specific pedestrian and carries richer fine-grained appearance features of pedestrians. To address the challenges in text-pedestrian image retrieval, this paper proposes an effective method that focuses on deep discriminative feature mining and suppression of sample diversity.
\subsection{Text-Pedestrian Image Retrieval}
In 2017, Li $et \ al$. \cite{17(0)} first introduced the concept of text-pedestrian image retrieval and constructed a large-scale benchmark dataset called CUHK-PEDES for this task. Subsequently, this task has received widespread attention, and a series of effective methods have emerged. Based on the characteristics of these methods, they can be broadly categorized into three classes: word-level relationship construction (WRC)-based methods \cite{1(7),2(8),3(10),4(12),5(18)}, cross-modal semantic alignment (CSA)-based methods \cite{6(4),7(11),8(37),9(42),10(47),11(57)}, and external model assistance (EMA)-based methods \cite{12(14),13(19),14(29),15(27),16(32)}.

The WRC-based methods often establish correspondences between words or phrases and local regions in image by measuring their similarity. Specifically, Niu $et \ al$. \cite{1(7)} established correspondences at the global-global, global-local, and local-local between image and text. Gao $et \ al$.\cite{2(8)} calculated the similarity between image and text features at three different scales: global-sentence, region-phrase, and patch-word, to construct correspondences between cross-modal full-scale features. In terms of cross-modal semantic alignment (CSA), Yan $et \ al$. \cite{8(37)} constructed a set of modality-shared semantic topic centers, and used the affinity between topic centers and features to achieve semantic alignment between text and image. Li $et \ al$. \cite{9(42)} proposed using local features of image to guide the extraction of local features from text to ensure that they have consistent semantics. Yang $et \ al$. \cite{11(57)} used visual and text features to predict the masked words in a text, thereby achieving semantic alignment of global features. The EMA-based methods typically leverage external models to assist in extracting features from text and image. Specifically, Wang $et \ al$. \cite{13(19)} utilized the guidance of a human parsing model to achieve alignment between local image and text features. Jing $et \ al$. \cite{12(14)} used a human pose estimation model to align different body parts with their corresponding textual descriptions. Aggarwal $et \ al$. \cite{15(27)} mined attribute annotations from a text corpus using NLTK \cite{25(65)} and used attribute annotations to bridge the modality gap between image and text, and facilitate representation learning. However, these methods rely too much on external models, making them less suitable for deployment and application in real-world scenarios.

Due to the stronger applicability of large-scale models, research on text-pedestrian image retrieval based on visual-language pre-training (VLP) has gained tremendous success lately. In particular, Yan $et \ al$. \cite{26(38)}, leveraging the advantages of the CLIP model \cite{27(2-5)} in multimodal feature representation, proposed a CLIP-driven fine-grained information mining framework. This method fully utilized the ability of CLIP to mine multimodal knowledge for text-image person retrieval. Jiang $et \ al$. \cite{10(47)} used the parameters of pre-trained CLIP to initialize the backbone of the text-pedestrian image retrieval model. On the other hand, Bai $et \ al$. \cite{28(56)} used the parameters of the pre-trained VLP model ALBEF \cite{29(2-19)} as the initial parameters of the backbone network to enable the model to extract discriminative features. Although methods based on large-scale models can significantly improve retrieval performance, they require higher hardware resources, which is not conducive to deployment in scenarios with limited computational and storage resources. In contrast to the above methods, the proposed method explores high-quality feature representation from two aspects: deep feature mining and suppression of interference caused by sample diversity. Meanwhile, our method does not rely on large-scale models to improve feature representation, it can achieve or even surpasses the performance achieved by the large-scale model-based method.
\section{The Proposed Method}
\subsection{Overview }
The overall network architecture of the proposed method is shown in Fig. \ref{label2}. It can be seen that the method consists of encoders, progressive feature mining(PFM), and external knowledge-assisted feature representation(EKFR). The encoders are composed of an image encoder and a text encoder, which are responsible for extracting the features of the input image and text respectively. PFM is used to mine discriminative features from neglected information, thereby avoiding the loss of discriminative information and improving the representation ability of features.
EKFR improves the discriminativeness of the current sample features by using other samples' knowledge.
\subsection{Feature Encoders}
In this paper, ViT \cite{31(63)} pre-trained on ImageNet \cite{30(59)} is used as the image encoder, denoted as $\bm E_{V}$. Given an image $\bm x_{v_{j}}$, it is divided into $n$ equally-sized and non-overlapping image patches. Each patch is vectorized and then undergoes a linear mapping to obtain $\tilde{\bm f}_{v_{j},i} \in \mathbb R^{d\times1}(i=1,2,\cdots,n)$. The entire image can be represented as $\{\tilde{\bm f}_{v_{j},1},\tilde{ \bm f}_{v_{j},2},\cdots,\tilde{\bm f}_{v_{j},n}\}$. Subsequently, the image representation is concatenated with a class token $\tilde{ \bm f}_{v_{j},c} \in \mathbb R^{d\times1}$ to form $ \tilde{\bm F}_{v_{j}}=\{\tilde{\bm f}_{v_{j},c};\tilde{\bm f}_{v_{j},1},\tilde{\bm f}_{v_{j},2}\cdots,\tilde{\bm f}_{v_{j},n}\} \in \mathbb R^{d\times (n+1)}$, which is fed into the ViT. The output of the image encoder is represented as $ \bm F_{v_{j}}=\{\bm f_{v_{j},c}; \bm f_{v_{j},1}, \bm f_{v_{j},2}\cdots, \bm f_{v_{j},n}\} \in \mathbb R^{d\times (n+1)}$.

For text feature extraction, the pre-trained BERT \cite{32(64)}model is employed as the text encoder, denoted as $\bm E_{B}$. Given a text $\bm x_{t_{j}}$, each word is first encoded as a one-hot vector and then undergoes word embedding to obtain the text representation, the representation of the $ i$-th word is $\tilde{\bm f}_{t_{j},i} \in \mathbb R^{d\times 1}(i=1,2,\cdots,m)$. The text representation is padded with class token $\tilde{\bm f}_{t_{j},c} $ at the beginning, represented as $ \tilde{\bm F}_{t_{j}}=\{\tilde{\bm f}_{t_{j},c};\tilde{\bm f}_{t_{j},1},\tilde{\bm f}_{t_{j},2}\cdots,\tilde{\bm f}_{t_{j},m}\} \in \mathbb R^{d\times (m+1)}$. The text feature representation is fed into BERT to generate text features $ \bm F_{t_{j}}=\{\bm f_{t_{j},c}; \bm f_{t_{j},1}, \bm f_{t_{j},2}\cdots, \bm f_{t_{j},m}\} \in \mathbb R^{d\times (m+1)} $.

To ensure that the features $ \bm F_{v_{j}}=\{\bm f_{v_{j},c}; \bm f_{v_{j},1}, \bm f_{v_{j},2}\cdots,\bm f_{v_{j},n}\} $ and $ \bm F_{t_{j}}=\{\bm f_{t_{j},c}; \bm f_{t_{j},1}, f_{t_{j},2}\cdots, \bm f_{t_{j},m}\} $ possess strong discriminability, we employ the cross-modal projection matching (CMPM) and cross-modal projection classification (CMPC)\cite{33(49)} loss functions to optimize $\bm E_{V}$ and $\bm E_{B}$ :
\begin{equation}
\begin{aligned}
    \bm L_{cm1}=(\bm L_{v2t}+\bm L_{t2v})+(\bm L_{vpt}+\bm L_{tpv})
\end{aligned},
\end{equation}
where $\bm L_{v2t}$ and $\bm L_{vpt}$ are the CMPM and CMPC losses of image-text direction, $\bm L_{t2v}$ and $\bm L_{tpv}$ are the CMPM and CMPC losses of text-image direction, respectively.
\subsection{Progressive Feature Mining}
In cross-modal retrieval tasks, modality difference is a key factor that hinders cross-modal matching. Among existing methods, extracting shared discriminative features from multi-modal sample pairs is one of the main approaches to address modality differences. However, since the undesigned feature extraction network tends to extract the most discriminative features locally, it reduces the robustness of the model. In order to enable the feature extraction network to pay attention to the more discriminative information of the input image and text, this paper proposes a PFM method, which enables the feature extraction network to focus on previously ignored discriminative features through a progressive boosting mechanism.

\begin{figure}[t!]
\centering
\includegraphics[width=2.7in,height=2.8in]{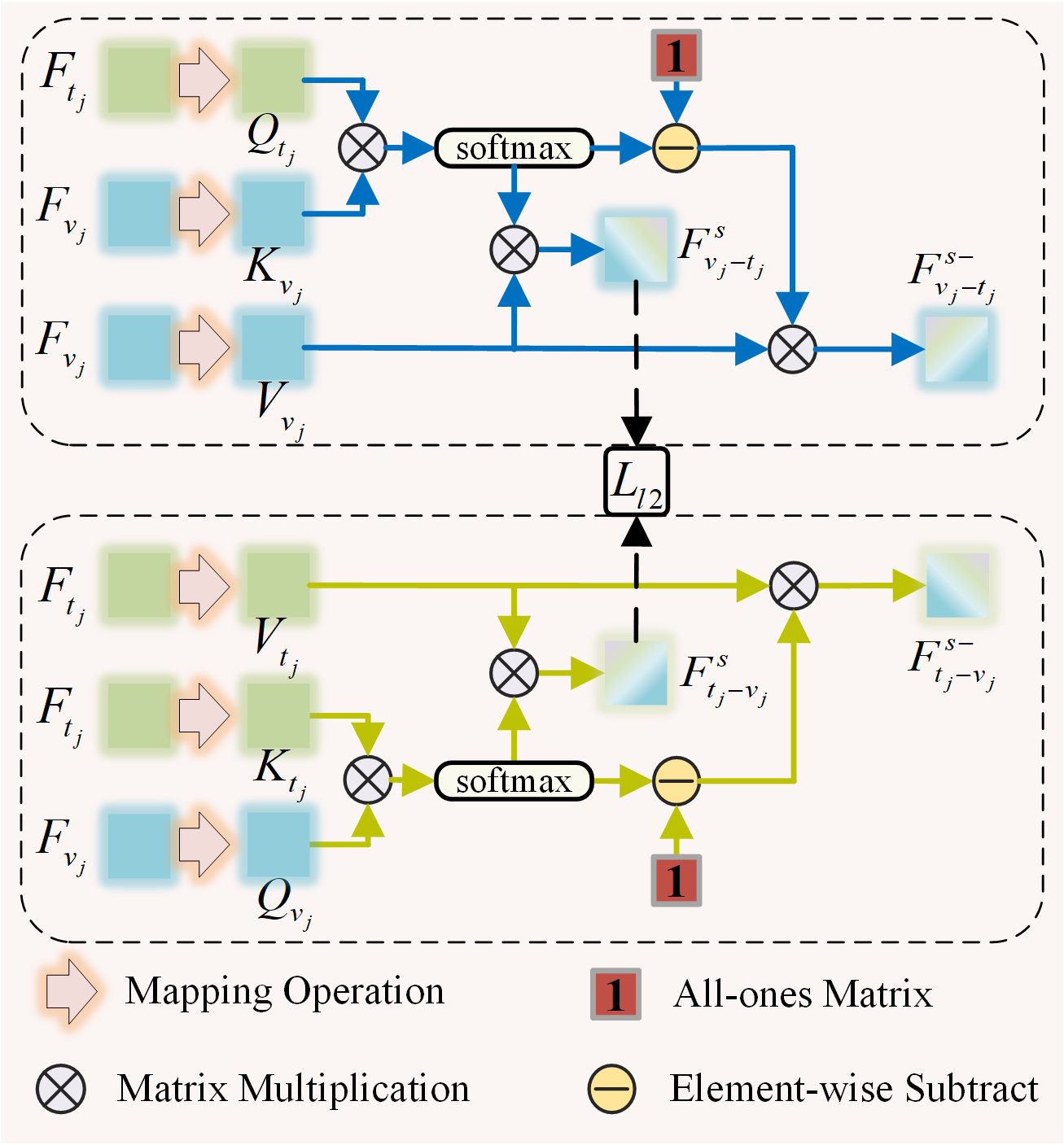}
\caption{Illustration of the dual reverse attention module (DRAM), which is in the stage \uppercase\expandafter{\romannumeral1} of PFM.}
\label{label3}
\end{figure}

The method consists of two stages of progressive feature mining, as illustrated in  Fig. \ref{label2}. To encourage $\bm E_{V}$ and $\bm E_{B}$ to focus on features beyond shared information and enhance their ability to extract discriminative features, this paper designs a 
dual reverse attention module (DRAM) in the stage \uppercase\expandafter{\romannumeral1} of PFM, the specific process of this module is depicted in Fig. \ref{label3}.
The method takes $ \bm F_{v_{j}}=\{\bm f_{v_{j},c}; \bm f_{v_{j},1}, \bm f_{v_{j},2}\cdots, \bm f_{v_{j},n}\} $  and $ \bm F_{t_{j}}=\{\bm f_{t_{j},c}; \bm f_{t_{j},1}, \bm f_{t_{j},2}\cdots, \bm f_{t_{j},m}\} $ as inputs and applies linear mapping to obtain $\bm Q_{v_{j}}=\bm W_{Q,v}\bm F_{v_{j}}$, $\bm K_{v_{j}}=\bm W_{K,v}\bm F_{v_{j}}$, $\bm V_{v_{j}}=\bm W_{V,v}\bm F_{v_{j}}$ and $\bm Q_{t_{j}}=\bm W_{Q,t}\bm F_{t_{j}}$, $\bm K_{t_{j}}=\bm W_{K,t}\bm F_{t_{j}}$, $\bm V_{t_{j}}=\bm W_{V,t}\bm F_{t_{j}}$. The cross-modal cross-attention between $\bm F_{v_{j}}$ and $\bm F_{t_{j}}$ can be represented as:
\begin{equation}
\begin{aligned}
    \bm F_{v_{j}-t_{j}}^{s}=softmax\bigg( \frac{\bm Q_{t_{j}}(\bm K_{v_{j}})^{T}}{\sqrt{d}} \bigg)\bm V_{v_{j}}
\end{aligned},
\end{equation}
\begin{equation}
\begin{aligned}
    \bm F_{t_{j}-v_{j}}^{s}=softmax\bigg( \frac{\bm Q_{v_{j}}(\bm K_{t_{j}})^{T}}{\sqrt{d}} \bigg)\bm V_{t_{j}}
\end{aligned}.
\end{equation}
The features $\bm F_{v_{j}-t_{j}}^{s}$ and $\bm F_{t_{j}-v_{j}}^{s}$ highlight the common information between $\bm F_{v_{j}}$ and $\bm F_{t_{j}}$, which positively contributes to cross-modal matching. The $\bm l2$ loss is employed to ensure the consistency of features $\bm F_{v_{j}-t_{j}}^{s}$ and $\bm F_{t_{j}-v_{j}}^{s}$:
\begin{equation}
\begin{aligned}
    \bm L_{l2}=\sum_{j=1}^N\bigg|\bigg| \bm Avgpool\big(\bm F_{v_{j}-t_{j}}^{s}\big)-\bm Avgpool\big(\bm F_{t_{j}-v_{j}}^{s}\big) \bigg|\bigg|_{2}
\end{aligned},
\end{equation}
where $\bm N$ and $\bm Avgpool$ denote the batch size and the average pooling, respectively.

However, to improve the matching ability of the model, it is far from enough to only rely on the information carried by $\bm F_{v_{j}-t_{j}}^{s}$  and $\bm F_{t_{j}-v_{j}}^{s}$. To solve this problem, this paper proposes to further extract discriminative information from features not attended by cross-attention. In this process, the information complementary to the information concerned by Eq.(2) and (3) can be denoted as:
\begin{equation}
\begin{aligned}
    \bm F_{v_{j}-t_{j}}^{s-} = \Bigg(\bm 1-softmax\bigg(\frac{\bm Q_{t_{j}}(\bm K_{v_{j}})^{T}}{\sqrt{d}}\bigg)\Bigg)\bm V_{v_{j}}
\end{aligned},
\end{equation}
\begin{equation}
\begin{aligned}
    \bm F_{t_{j}-v_{j}}^{s-} = \Bigg(\bm 1-softmax\bigg(\frac{\bm Q_{v_{j}}(\bm K_{t_{j}})^{T}}{\sqrt{d}}\bigg)\Bigg)\bm V_{t_{j}}
\end{aligned},
\end{equation}
where $\bm 1$ is an all-ones matrix.

In order to improve the ability of the model to mine discriminative features, we introduce new Transformer Encoder \cite{54} layers after the dual reverse attention and the encoders $\bm E_{V}$ and $\bm E_{B}$ respectively. To ensure that the newly introduced Transformer layers also have the ability to mine complementary information, the parameters of the two Transformer layers that extract specific modal features are shared. Let $\bm E_{v}$ and $\bm E_{t}$ represent the newly introduced Transformer layers for pedestrian image and text feature extraction, respectively,  with inputs $\bm F_{v_{j}-t_{j}}^{s-}$ and $\bm F_{t_{j}-v_{j}}^{s-}$. To enable $\bm E_{v}$ and $\bm E_{t}$ to further extract discriminative features from $\bm F_{v_{j}-t_{j}}^{s-}$ and $\bm F_{t_{j}-v_{j}}^{s-}$ to make up for the lack of discriminative information of  $\bm F_{v_{j}}$ and $\bm F_{t_{j}}$, we input $\bm F_{v_{j}-t_{j}}^{s-}$ and $\bm F_{t_{j}-v_{j}}^{s-}$ into $\bm E_{v}$ and $\bm E_{t}$ respectively to obtain the supplementary information $\bm F_{v_{j}}^{c}$ and $\bm F_{t_{j}}^{c}$ for $\bm F_{v_{j}}$ and $\bm F_{t_{j}}$:
\begin{equation}
\begin{aligned}
    \bm F_{v_{j}}^{c} = \bm E_{v}(\bm F_{v_{j}-t_{j}}^{s-})+\bm F_{v_{j}-t_{j}}^{s}
\end{aligned},
\end{equation}
\begin{equation}
\begin{aligned}
    \bm F_{t_{j}}^{c} = \bm E_{t}(\bm F_{t_{j}-v_{j}}^{s-})+\bm F_{t_{j}-v_{j}}^{s}
\end{aligned},
\end{equation}
where the incorporation of $\bm F_{v_{j}-t_{j}}^{s}$ and $\bm F_{t_{j}-v_{j}}^{s}$ is aimed at preventing the loss of previously extracted shared information. To ensure the discriminability of features $\bm F_{v_{j}}^{c}$ and $\bm F_{t_{j}}^{c}$, we utilize ID loss to optimize $\bm E_{V}$, $\bm E_{B}$, $\bm E_{v}$, $\bm E_{t}$ and the parameters in DRAM:
\begin{equation}
\begin{aligned}
    \bm L_{id1} = \sum_{j=1}^{N}\bm CE(\bm f_{v_{j},c}^c, \bm y_{v_{j}})+\bm CE(\bm f_{t_{j},c}^c, \bm y_{t_{j}})
\end{aligned},
\end{equation}
where $\bm f_{v_{j},c}^c$ and $\bm f_{t_{j},c}^c$ are the class token features of $\bm F_{v_{j}}^{c}$ and $\bm F_{t_{j}}^{c}$, $\bm y_{v_{j}}$ and $\bm y_{t_{j}}$ are the labels of the samples $\bm x_{v_{j}}$ and $\bm x_{t_{j}}$ respectively. $\bm CE(\cdot)$ is the cross-entropy loss function, $\bm N$ is the batchsize. The features $\bm F_{v_{j}}$ and $\bm F_{t_{j}}$ are fed into $\bm E_{v}$ and $\bm E_{t}$, and added to the original features $\bm F_{v_{j}}$ and $\bm F_{t_{j}}$, which can be represented as follows:
\begin{equation}
\begin{aligned}
    \bm F_{v_{j}}^{o} = \bm E_{v}(\bm F_{v_{j}})+\bm F_{v_{j}}
\end{aligned},
\end{equation}
\begin{equation}
\begin{aligned}
    \bm F_{t_{j}}^{o} = \bm E_{t}(\bm F_{t_{j}})+\bm F_{t_{j}}
\end{aligned}.
\end{equation}

\begin{figure}[t!]
\centering
\includegraphics[width=2.7in,height=2.8in]{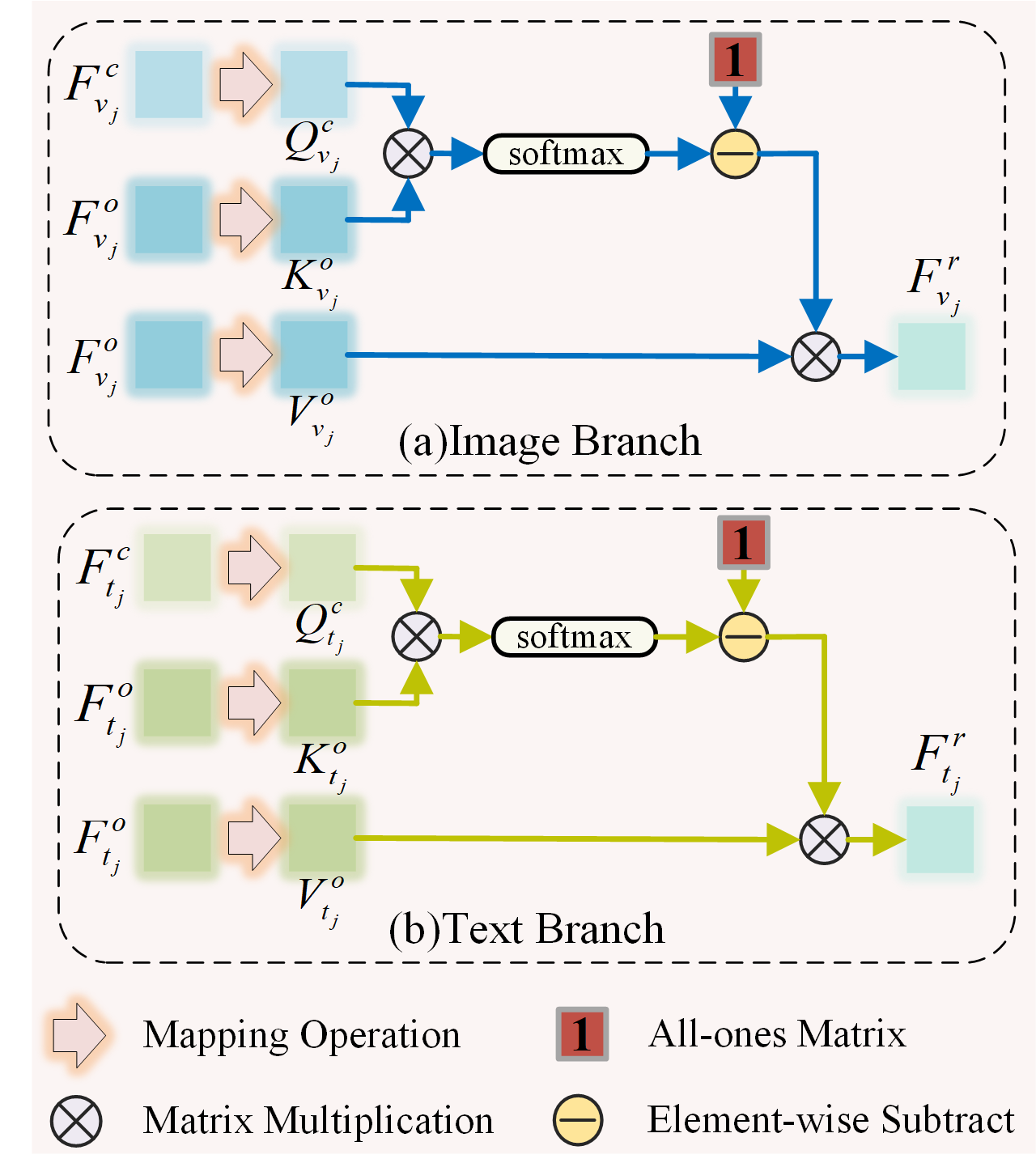}
\caption{Illustration of the single reverse attention module (SRAM), which is in the stage \uppercase\expandafter{\romannumeral2} of PFM.}
\label{label4}
\end{figure}

$\bm E_{v}$ and $\bm E_{t}$ are inclined to extract overlooked information from $\bm F_{v_{j}-t_{j}}^{s-}$ and $\bm F_{t_{j}-v_{j}}^{s-}$, respectively. When $\bm F_{v_{j}} (\bm F_{t_{j}})$ is fed into $\bm E_{v} (\bm E_{t})$, we expect the information extracted by $\bm E_{v} (\bm E_{t})$ from $\bm F_{v_{j}} (\bm F_{t_{j}})$ not only contains the information of $\bm F_{v_{j}}^{c} (\bm F_{t_{j}}^{c})$, but also other pedestrian-related information that is different from $\bm F_{v_{j}}^{c} (\bm F_{t_{j}}^{c})$, thereby achieving progressive discriminative feature mining. Thus, we propose a method called single reverse attention module (SRAM) in the stage \uppercase\expandafter{\romannumeral2} of PFM, aiming at extracting pedestrian-related features from $\bm F_{v_{j}}^{o} (\bm F_{t_{j}}^{o})$ that are inconsistent with $\bm F_{v_{j}}^{c} (\bm F_{t_{j}}^{c})$ but still relevant to pedestrian identity. The specific process is illustrated in Fig. \ref{label4}. Let $\bm F_{v_{j}}^{o} (\bm F_{t_{j}}^{o})$, $\bm F_{v_{j}}^{c} (\bm F_{t_{j}}^{c})$ undergo a linear mapping to obtain $\bm Q_{v_{j}}^{c}=\bm W_{Q,v}^{c}\bm F_{v_{j}}^{c} (\bm Q_{t_{j}}^{c}=\bm W_{Q,t}^{c}\bm F_{t_{j}}^{c})$, $\bm K_{v_{j}}^{o}=\bm W_{K,v}^{o}\bm F_{v_{j}}^{o} (\bm K_{t_{j}}^{o}=\bm W_{K,t}^{o}\bm F_{t_{j}}^{o})$, and $\bm V_{v_{j}}^{o}=\bm W_{V,v}^{o}\bm F_{v_{j}}^{o} (\bm V_{t_{j}}^{o}=\bm W_{V,t}^{o}\bm F_{t_{j}}^{o})$. After removing the relevant information from $\bm F_{v_{j}}^{o}$ and $\bm F_{v_{j}}^{c}$ as well as from $\bm F_{t_{j}}^{o}$ and $\bm F_{t_{j}}^{c}$, the outputs are denoted as:
\begin{equation}
\begin{aligned}
    \bm F_{v_{j}}^{r} = \Bigg(\bm 1-softmax\bigg(\frac{\bm Q_{v_{j}}^{c}(\bm K_{v_{j}}^{o})^{T}}{\sqrt{d}}\bigg)\Bigg)\bm V_{v_{j}}^{o}
\end{aligned},
\end{equation}
\begin{equation}
\begin{aligned}
    \bm F_{t_{j}}^{r} = \Bigg(\bm 1-softmax\bigg(\frac{\bm Q_{t_{j}}^{c}(\bm K_{t_{j}}^{o})^{T}}{\sqrt{d}}\bigg)\Bigg)\bm V_{t_{j}}^{o}
\end{aligned}.
\end{equation}

To continually enhance the discriminative information carried by features $\bm F_{v_{j}}$ and $\bm F_{t_{j}}$, the features $\bm F_{v_{j}}^{r}$ and $\bm F_{t_{j}}^{r}$ obtained after removing the correlated information, should be closely related to the identity of the pedestrian. Therefore, we introduce the ID loss to ensure that $\bm F_{v_{j}}^{r}$ and $\bm F_{t_{j}}^{r}$ exhibit strong classification capabilities. The ID loss optimizes  $\bm E_{V}$, $\bm E_{B}$, $\bm E_{v}$, $\bm E_{t}$, as well as the parameters in DRAM and SRAM:
\begin{equation}
\begin{aligned}
    \bm L_{id2} = \sum_{j=1}^{N}\bm CE(\bm f_{v_{j},c}^r, \bm y_{v_{j}})+\bm CE(\bm f_{t_{j},c}^r, \bm y_{t_{j}})
\end{aligned},
\end{equation}
where $\bm f_{v_{j},c}^r$ and $\bm f_{t_{j},c}^r$ denote the class token features of $\bm F_{v_{j}}^{r}$ and $\bm F_{t_{j}}^{r}$, respectively.
\begin{figure*}[t!]
\centering
\includegraphics[width=6.3in,height=3.2in]{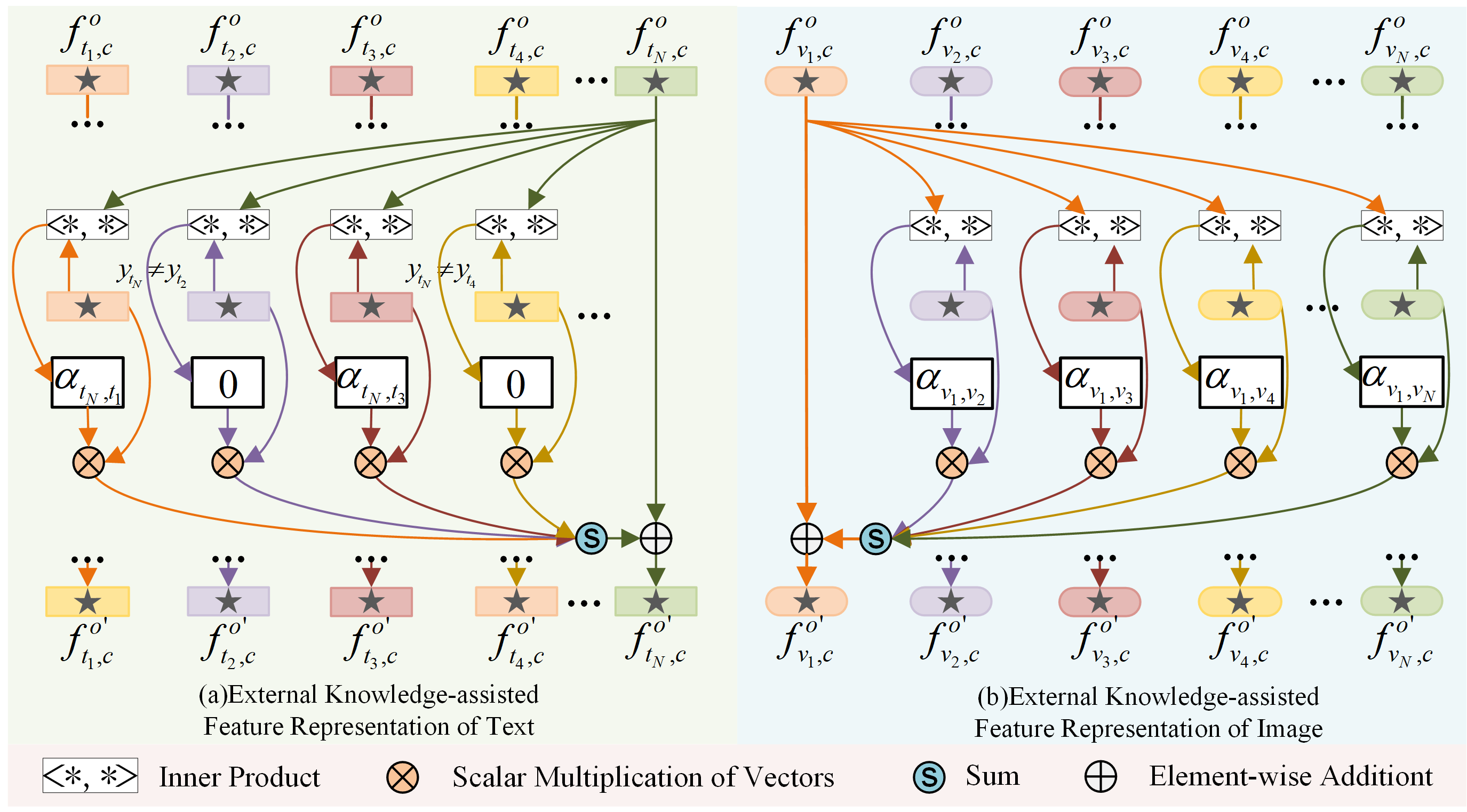}
\caption{External Knowledge-assisted Feature Representation. The external knowledge-assisted module leverages the features of other samples to assist in representing the features of the current sample, further mitigating modality disparities and enhancing the discriminative representation of the features.}
\label{label5}
\end{figure*}
\subsection{External Knowledge-assisted Feature Representation}
In text-pedestrian image retrieval, multiple texts describing the same pedestrian are usually available from different eyewitnesses. In this process, although it is clear that these eyewitnesses are describing the same pedestrian, due to the varying language styles and observed perspectives of the eyewitnesses, the texts describing the same pedestrian will be quite different. Effectively suppressing the differences among samples of the same identity, while synthesizing the complementary information brought about by the perspectives differences, is crucial for cross-modal identity matching. Similar issues also exist in pedestrian images, where images of the same pedestrian often show differences and sameness.

To address this, we propose an External Knowledge-assisted Feature Representation (EKFR) method. This method leverages the information from other samples with the same modality to enhance the identity-related information and suppress irrelevant information. The process of this method is illustrated in Fig. \ref{label5}. In the external knowledge-assisted feature representation for text modality, we take the class token $\bm f_{t_{j}, c}^{o}$  from the features $\bm F_{t_{j}}^{o}$ as input, the cosine similarity between $\bm f_{t_{j},c}^{o}$ and another sample's feature $\bm f_{t_{i},c}^{o}$ is calculated as:
\begin{equation}
\begin{aligned}
    s_{t_{j},t_{i}}=\frac{(\bm W_{Q}\bm f_{t_{j},c}^{o})^{T}(\bm W_{K}\bm f_{t_{i},c}^{o})}{\|\bm W_{Q}\bm f_{t_{j},c}^{o}\|\|\bm W_{K}\bm f_{t_{i},c}^{o}\|}
\end{aligned},
\end{equation}
where $\bm W_{Q} \in \mathbb R^{d\times d}$ and $\bm W_{K} \in \mathbb R^{d\times d}$ are two linear mapping matrices.
If two samples belong to the same identity, the weight is the result of cosine similarity after softmax. If two samples do not belong to the same identity, the weight is $0$. This process can be formalized as:
\begin{equation}
   \alpha_{t_{j},t_{i}}=
   \begin{cases}
   \frac{\exp (s_{t_{j},t_{i}})}{\sum_{i=1}^N\exp(s_{t_{j},t_{i}})}, & \text{if}~~ \bm y_{t_{j}}=\bm y_{t_{i}}\\
   0, & \text{if}~~ \bm y_{t_{j}}\ne \bm y_{t_{i}}
\end{cases},
\end{equation}
where $\bm y_{t_{j}}$ and $\bm y_{t_{i}}$ are the labels of $\bm x_{t_{j}}$ and $\bm x_{t_{i}}$, respectively. With the weight $\alpha_{t_{j},t_{i}}(i\in \{1,2,3,\cdots N\},i\ne j)$, we can aggregate all samples in the batch, and the result after information supplementation can be formulated as:
\begin{equation}
\begin{aligned}
    \bm f_{t_{j},c}^{o^\prime}=\sum_{i=1,i \ne j}^{N}\alpha_{t_{j},t_{i}}(\bm W_{V}\bm f_{t_{i},c}^{o})+\bm f_{t_{j},c}^{o}
\end{aligned},
\end{equation}
where $\bm W_{V} \in \mathbb R^{d\times d}$ is a linear mapping matrix.

Since the weight is 0 when two samples do not belong to the same identity, only sample features of the same identity are used to enhance the feature representation of the current sample in the case of text features. During this process, features consistent with the current sample information are enhanced, while features inconsistent with the current sample information are suppressed to a certain extent. At the same time, since the strategy of weighted fusion is adopted to realize the feature representation, it realizes the synthesis of complementary information.

In the external knowledge-assisted feature representation for image modality, we take the class token $\bm f_{v_{j},c}^{o}$  from the features $\bm F_{v_{j}}^{o}$ as input, and calculate the cosine similarity between $\bm f_{v_{j},c}^{o}$ and another sample's feature $\bm f_{v_{i},c}^{o}$:
\begin{equation}
\begin{aligned}
    s_{v_{j},v_{i}}=\frac{(\bm W_{Q}\bm f_{v_{j},c}^{o})^{T}(\bm W_{K}\bm f_{v_{i},c}^{o})}{\|\bm W_{Q}\bm f_{v_{j},c}^{o}\|\|\bm W_{K}\bm f_{v_{i},c}^{o}\|}
\end{aligned},
\end{equation}
where $\bm W_{Q} \in \mathbb R^{d\times d}$ and $\bm W_{K} \in \mathbb R^{d\times d}$ are two linear mapping matrices.
The cosine similarity after softmax can be expressed as:
\begin{equation}
\begin{aligned}
    \alpha_{v_{j},v_{i}}= \frac{\exp (s_{v_{j},v_{i}})}{\sum_{i=1}^N\exp(s_{v_{j},v_{i}})}
\end{aligned}.
\end{equation}
Aggregating all samples in the same batch using weights $\alpha_{v_{j},v_{i}}(i\in \{1,2,3,\cdots N\},i\ne j)$, the result after information supplementation can be denoted as $\bm f_{v_{j},c}^{o^\prime}$:
\begin{equation}
\begin{aligned}
    \bm f_{v_{j},c}^{o^\prime}=\sum_{i=1,i \ne j}^{N}\alpha_{v_{j},v_{i}}(\bm W_{V}\bm f_{v_{i},c}^{o})+\bm f_{v_{j},c}^{o}
\end{aligned},
\end{equation}
where $\bm W_{V} \in \mathbb R^{d\times d}$ is a linear mapping matrix.

In the aforementioned process, to learn the shared features across modalities, we utilize modality-shared projection matrices to project the class token of each modality. To ensure the consistency between the image features and text features with the same identity after external knowledge assistance, for output features $\bm f_{v_{j},c}^{o^\prime}$ and $\bm f_{t_{j},c}^{o^\prime}$, we use CMPM and CMPC losses to optimize $\bm E_{V}$ and $\bm E_{B}$, $\bm E_{v}$, $\bm E_{t}$, as well as the parameters in PFM and EKFR:
\begin{equation}
\begin{aligned}
    \bm L_{cm2}=\bm L_{cmpm2}+\bm L_{cmpc2}
\end{aligned},
\end{equation}
where $\bm L_{cmpm2}$ and $\bm L_{cmpc2}$ have the same form as Eq.(1).

In the aforementioned method, each current pedestrian image interacts with all other samples in a batch. When two images belong to the same identity, this approach effectively emphasizes the consistent information relevant to identity clues, while suppressing the influence of identity-irrelevant information. Conversely, when two images are from different pedestrian identities, the method efficiently suppresses the interference of similar features, thereby enhancing the impact of identity-related features. For textual descriptions, this study only conducts intra-class knowledge-assisted feature representation among samples of the same identity. This is mainly because the text descriptions are manually constructed, and texts describing the same pedestrian can be known in advance. Therefore, using the text knowledge describing the same pedestrian to assist the representation of the current text can not only effectively suppress the impact of sentence style differences, but also effectively highlight the role of useful information.
\subsection{Model Optimization and Training}
The entire network is trained in an end-to-end manner, and the model parameters are optimized by the following total loss:
\begin{equation}
\begin{aligned}
    \bm L=\bm L_{cm}+\lambda_{1}\bm L_{id}+\lambda_{2}\bm L_{l2}
\end{aligned},
\end{equation}
where $\bm L_{cm}=\bm L_{cm1}+\bm L_{cm2}$, $\bm L_{id}=\bm L_{id1}+\bm L_{id2}$, $\lambda_{1}$ and $\lambda_{2}$ are hyperparameters used to balance the contributions of the $\bm L_{id}$ and $\bm L_{l2}$. In order to better understand the proposed method, we provide the training process of the model in detail in \textbf{Algorithm 1}.
\begin{algorithm}[!t]\small
 \caption{ Progressive Feature Mining and External Knowledge-assisted Text-Pedestrian Image Retrieval}\label{alg:A}
\begin{algorithmic}
\STATE {\textbf{Input:} Image set $\bm X_{v}=\{\bm x_{v_{i}}\}_{i=1}^{N}$, text set $\bm X_{t}=\{\bm x_{t_{i}}\}_{i=1}^{N}$, the corresponding
pedestrian identity labels $\bm Y_{v}=\{\bm y_{v_{i}}\}_{i=1}^{N}$, $\bm Y_{t}=\{\bm y_{t_{i}}\}_{i=1}^{N}$, pre-trained models ViT, BERT, hyperparameters $\lambda_{1}$ and $\lambda_{2}$, number of iterations $T$.\\}
\STATE {\textbf{Output:} The trained encoder $\bm E_{V}$,$\bm E_{B}$,$\bm E_{v}$,$\bm E_{t}$,$\bm W_{Q}$,$\bm W_{K}$,$\bm W_{V}$.\\
\begin{flushleft}
~1:Sample a batch of labeled source data.\\
~2:Initialize $\bm E_{v}$,$\bm E_{t}$,$\bm W_{Q}$,$\bm W_{K}$,$\bm W_{V}$\\
~3:\textbf{for} \emph{iter}=1, $\cdots$, \emph{T} \textbf{do}\\
~4:\qquad Update $\bm E_{V}$,$\bm E_{B}$ by minimizing the loss in Eq.(1).\\
~5:\qquad Update $\bm E_{V}$,$\bm E_{B}$  by minimizing the loss in Eq.(4).\\
~6:\qquad Update $\bm E_{V}$,$\bm E_{B}$,$\bm E_{v}$,$\bm E_{t}$ by minimizing the loss in Eq.(9).\\
~7:\qquad Update $\bm E_{V}$,$\bm E_{B}$,$\bm E_{v}$,$\bm E_{t}$ by minimizing the loss in Eq.(14).\\
~8:\qquad Update $\bm E_{V}$,$\bm E_{B}$,$\bm E_{v}$,$\bm E_{t}$, $\bm
          W_{Q}$,$\bm W_{K}$,$\bm W_{V}$ by minimizing the \\
    \qquad       loss in Eq.(21).\\
~9:\textbf{end for}\\
\end{flushleft}}
\end{algorithmic}
\end{algorithm}

\section{Experiments}
\subsection{Datasets and Evaluation Protocol}
\textbf{Datasets}. We evaluated the proposed method on three commonly used text-pedestrian image retrieval datasets, namely CUHK-PEDES dataset\cite{17(0)}, ICFG-PEDES dataset\cite{3(10)}, and RSTPReid dataset\cite{34(25)}.

\textbf{CUHK-PEDES dataset} is a widely used large-scale benchmark dataset for text-based pedestrian image retrieval. The dataset contains 40,206 images and 80,412 text descriptions of 13,003 pedestrians. Each image has two text descriptions manually labeled, and the average length of each text description is no less than 23 words. Following the protocol in \cite{17(0)}, 34,054 images of 11,003 individuals and corresponding 68,108 text descriptions are used as the training set. The validation set contains 6,156 text descriptions and 3,078 person images of 1,000 identities. The test set contains 6,148 text descriptions and 3,074 person images of the remaining 1,000 pedestrians.

\textbf{ICFG-PEDES dataset} is a newly collected dataset that contains more detailed descriptions of pedestrians. The dataset contains a total of 54,522 images and 54,522 text descriptions of 4,102 pedestrians, and the average length of each text description is 37 words. According to the division rules of the dataset\cite{3(10)}, the dataset is divided into a training set and a test set, the training set contains 34,674 image-text pairs of 3102 pedestrians, and the test set contains 19,848 image-text pairs of 1,000 pedestrians.

\begin{table}[!ht]\small
\centering {\caption{Comparison of the proposed method with state-of-the-art methods on CUHK-PEDES dataset. The evaluation results on Rank-1, Rank-5, Rank-10, and mAP are listed. ``--'' indicates no reported data. The bolded data represents the optimal result.}\label{Table1}
\renewcommand\arraystretch{1.4}
\resizebox{\linewidth}{!}{
\begin{tabular}{c|c|c|c|c|c|c}
\hline\hline
Types&Methods&Ref&Rank-1 &Rank-5&Rank-10&mAP\\
\hline
\multirow{26}*{{\rotatebox{90}{w/o VLP}}}&MCCL \cite{39(50)}&ICASSP19&50.58&--&79.06&--\\
  &A-GANet \cite{40(26)}&ACM MM19&53.14&74.03&81.95&--\\
  &TIMAM \cite{41(9)}&ICCV19&54.51&77.56&84.78&--\\
  &Dual Path \cite{38(48)}&TOMM20&44.40&66.26&75.07&-\\
  &AATE \cite{62}&TMM20&52.42&74.98&82.74&-\\
  &MIA \cite{1(7)}&TIP20&53.10&75.00&82.90&--\\
  &PMA \cite{14(29)}&AAAI20&53.811&73.54&81.23&--\\
  &ViTTA \cite{13(19)}&ECCV20&55.97&75.84&83.52&--\\
  &IMG-Net \cite{43(62)}&JEI20&56.48&76.89&85.01&--\\
  &CMAAM \cite{15(27)}&WACV20&56.68&77.18&84.86&--\\
  &HGAN \cite{44(30)}&ACM MM20&59.00&79.49&86.62&--\\
  &CMKA \cite{42(51)}&TIP21&54.69&73.65&81.86&--\\
  &NAFS \cite{2(8)}&arXiv21&59.94&79.86&86.70&54.07\\
  &DSSL \cite{34(25)}&ACM MM21&59.98&80.41&87.56&--\\
  &MGEL \cite{46(1)}&IJCAI21&60.27&80.01&85.74&--\\
  &SSAN \cite{3(10)}&arXiv21&61.37&80.15&86.73&--\\
  &LapsCore \cite{47(20)}&ICCV21&63.40&-&87.80&--\\
  &SUM \cite{45(35)}&KBS22&59.22&80.35&87.60&--\\
  &ACSA \cite{61}&TMM22&63.56&81.40&87.70&--\\
  &ISANet \cite{8(37)}&arXiv22&63.92&82.15&87.69&--\\
  &LBUL \cite{48(24)}&ACM MM22&64.04&82.66&87.22&--\\
  &SAF \cite {6(4)}&ICASSP22&64.13&82.62&88.40&58.61\\
  &TIPCB \cite{4(12)}&Neuro22&64.26&83.19&89.10&--\\
  &CAIBC \cite{50(23)}&ACM MM22&64.43&82.87&88.37&--\\
  &AXM-Net \cite{51(21)}&AAAI22&64.44&80.52&86.77&58.73\\
  &LGUR \cite{7(11)}&ACM MM22&65.25&83.12&89.00&--\\
  &IVT \cite{52(22)}&ECCVW22&65.59&83.11&89.21&--\\
  &APTM \cite{11(57)}&ACM MM23&76.17&89.47&93.57&65.52\\
  &\textbf{PFM-EKFP}&This paper&\bf 77.24&\bf 93.71&\bf 96.98&\bf 73.47\\
\hline
\multirow{4}*{{\rotatebox{90}{w/ VLP}}}&TextReID \cite{49(13)}&BMVC21 &64.08&81.73&88.19&60.08\\
  &CFine \cite{26(38)}&arXiv22&69.57&85.93&91.15&--\\
  &IRRA \cite{10(47)}&CVPR23&73.38&89.93&93.71&66.13\\
  &RaSa \cite{28(56)}&arXiv23&76.51&90.29&94.25&69.38\\
\hline\hline
\end{tabular}}}
\end{table}

\textbf{RSTPReid dataset} is a dataset specially designed to handle text-to-pedestrian image retrieval tasks in real-world scenarios. The dataset contains 4,101 pedestrians with different identities, with a total of 20,505 images. Each pedestrian has 5 corresponding images taken by different cameras. Each image has two text descriptions, each no less than 23 words in length. Following \cite{34(25)}, the training set, validation set and test set contain 3,701, 200 and 200 pedestrians with different identities respectively, and the corresponding images are 18,505, 1,000 and 1,000, and the corresponding numbers of textual descriptions are 37,010, 2,000, and 2,000, respectively.

\textbf{Evaluation Protocol}. For performance evaluation, the Cumulative Match Characteristic (CMC), i.e., Rank-K  (higher values indicate better performance) is utilized as the metric to measure the retrieval performance of different methods. The Rank-K metric represents the model's ability to correctly match the query sample within the top-K retrieved results under a given query condition. The Rank-1, Rank-5, and Rank-10 accuracies are reported during the evaluation process. Additionally, to comprehensively assess the retrieval performance, mean Average Precision (mAP) is employed as another retrieval criterion. Using these metrics, we can objectively evaluate the retrieval performance of different methods on the three datasets.
\subsection{Implementation Details}
The proposed method use the ViT pre-trained on ImageNet \cite{30(59)} as the image feature encoder and the pre-trained BERT as the text feature encoder. Random horizontal flipping is used for data augmentation and all images are resized to 224$\times$224. Additionally, the maximum text length for input sentences is set to 100. After tokenization, the lengths of visual and text token sequences are 196 and 100, respectively, and the dimensions of image and text embedding are both set to 768. For training, we use the Adam optimizer \cite{35(60)} with a learning rate of 0.00001 for the image and text encoders and 0.0001 for other modules, the learning rate is adjusted according to the warm-up strategy \cite{36(61)}. The batch size is set to 32 during the training, and a total of 60 epochs are iterated. The learning rate decays with a factor of 0.1 at the 20th, 30th, and 40th epochs. The experiments are implemented on PyTorch library and conducted on a desktop computer with a single RTX 3090 GPU.
\begin{table}[!ht]\small
\centering {\caption{Comparison of the proposed method with state-of-the-art methods on ICFG-PEDES dataset. The evaluation results on Rank-1, Rank-5, Rank-10, and mAP are listed. ``--'' indicates no reported data. The bolded data represents the optimal result.}\label{Table2}
\renewcommand\arraystretch{1.4}
\resizebox{\linewidth}{!}{
\begin{tabular}{c|c|c|c|c|c|c}
\hline\hline
Types&Methods&Ref&Rank-1 &Rank-5&Rank-10&mAP\\
\hline
\multirow{8}*{{\rotatebox{90}{w/o VLP}}}&Dual Path \cite{38(48)}&TOMM20&38.99&59.44&68.41&--\\
  &MIA \cite{1(7)}&TIP20&46.49&67.14&75.18&--\\
  &ViTTA \cite{13(19)}&ECCV20&50.98&68.79&75.78&--\\
  &SSAN \cite{3(10)}&arXiv21&54.23&72.63&79.53&--\\
  &TIPCB \cite{4(12)}&Neuro22&54.96&74.72&81.89&--\\
  &IVT \cite{52(22)}&ECCVW22&56.04&73.60&80.22&--\\
  &ISANet \cite{8(37)}&arXiv22&57.73&75.42&81.72&--\\
  &APTM \cite{11(57)}&ACM MM23&68.22&82.87&87.50&39.58\\
  &\textbf{PFM-EKFP}&This paper&\bf 69.29&\bf 89.10&\bf 94.06&\bf 47.15\\
\hline
\multirow{3}*{{\rotatebox{90}{w/ VLP}}}&CFine \cite{26(38)}&arXiv22&60.83&76.55&82.42&--\\
  &IRRA \cite{10(47)}&CVPR23&63.46&80.25&85.82&38.06\\
  &RaSa \cite{28(56)}&arXiv23&65.28&80.40&85.12&41.29\\
\hline\hline
\end{tabular}}}
\end{table}
\begin{table}[!ht]\small
\centering {\caption{Comparison of the proposed method with state-of-the-art methods on RSTPReid dataset. Rank-1, Rank-5, Rank-10, and mAP are listed. ``--'' indicates no reported data. Bold data indicates the best among similar methods.}\label{Table3}
\renewcommand\arraystretch{1.4}
\resizebox{\linewidth}{!}{
\begin{tabular}{c|c|c|c|c|c|c}
\hline\hline
Types&Methods&Ref&Rank-1 &Rank-5&Rank-10&mAP\\
\hline
\multirow{8}*{{\rotatebox{90}{w/o VLP}}}&IMG-Net \cite{43(62)}&JEI20&37.60&61.15&73.55&--\\
  &AMEN \cite{53(66)}&PRCV21&38.45&62.40&73.80&--\\
  &DSSL \cite{34(25)}&ACM MM21&39.05&62.60&73.95&--\\
  &SSAN \cite{3(10)}&arXiv21&43.50&67.80&77.15&--\\
  &SUM \cite{45(35)}&KBS22&41.38&67.48&76.48&--\\
  &LBUL \cite{48(24)}&ACM MM22&45.55&68.20&77.85&--\\
  &IVT \cite{52(22)}&ECCVW22&46.70&70.00&78.80&--\\
  &ACSA \cite{61}&TMM22&48.40&71.85&81.45&--\\
  &\textbf{PFM-EKFP}&This paper&\bf 48.65&\bf 77.15&\bf 87.00&\bf 43.08\\
\hline
\multirow{3}*{{\rotatebox{90}{w/ VLP}}}&CFine \cite{26(38)}&arXiv22&50.55&72.50&81.60&--\\
  &IRRA \cite{10(47)}&CVPR23&60.20&81.30&88.20&47.17\\
  &RaSa \cite{28(56)}&arXiv23&66.90&86.50&91.35&52.31\\
\hline\hline
\end{tabular}}}
\end{table}

\subsection{Comparison with State-of-the-Art Methods}
In this section, we evaluate the proposed method by comparing with the state-of-the-art methods on three public databases, as shown in Tables \ref{Table1}--\ref{Table3}. The proposed method demonstrates superior retrieval performance on all three datasets.

\textbf{Results on CUHK-PEDES dataset}: We validate the proposed method on CUHK-PEDES benchmark dataset and compare it with other methods. The compared methods include MCCL \cite{39(50)}, A-GANet \cite{40(26)}, TIMAM \cite{41(9)}, Dual Path \cite{38(48)}, AATE \cite{62}, MIA \cite{1(7)}, PMA \cite{14(29)}, ViTTA \cite{13(19)}, IMG-Net \cite{43(62)}, CMAAM \cite{15(27)}, HGAN \cite{44(30)}, CMKA \cite{42(51)}, NAFS \cite{2(8)}, DSSL \cite{34(25)}, MGEL \cite{46(1)}, SSAN \cite{3(10)}, LapsCore \cite{47(20)}, SUM \cite{45(35)}, ACSA \cite{61}, ISANet \cite{8(37)}, LBUL \cite{48(24)}, SAF \cite{6(4)}, TIPCB \cite{3(10)}, CAIBC \cite{50(23)}, AXM-Net \cite{51(21)}, LGUR \cite{7(11)}, IVT \cite{52(22)}, APTM \cite{11(57)}, TextReID \cite{49(13)}, CFine \cite{26(38)}, IRRA \cite{10(47)} and RaSa \cite{28(56)}. The comparison results are shown in Table \ref{Table1}, it can be observed that our proposed method outperforms the compared methods. It achieves 77.24\%, 93.71\%, and 96.98\% accuracy at Rank-1, Rank-5, and Rank-10, respectively, and 0.73\%, 3.42\% and 2.73\% improvement on the Rank-1, Rank-5 and Rank-10 are obtained over the suboptimal approach RaSa \cite{28(56)}. Additionally, our method achieves a mAP value of 73.48\%, which is 4.10\% higher than that of RaSa. It is noteworthy that the methods CFine \cite{26(38)}, IRRA \cite{10(47)}, and RaSa \cite{28(56)} with high-performance utilize large-scale VLP models such as CLIP and ALBEF as the backbones to extract features, while our proposed method does not utilize such VLP models. Despite this, the proposed method surpasses these approaches, demonstrating the effectiveness of the proposed method.

\textbf{Results on ICFG-PEDES dataset}: To further demonstrate the effectiveness of the proposed method, we conduct experiments on ICFG-PEDES dataset. We compared our method with the latest 11 methods, including Dual Path \cite{38(48)}, MIA \cite{1(7)}, ViTTA \cite{13(19)}, SSAN \cite{3(10)}, TIPCB \cite{3(10)}, IVT \cite{52(22)}, ISANet \cite{8(37)}, APTM \cite{11(57)}, CFine \cite{26(38)}, IRRA \cite{10(47)} and RaSa \cite{28(56)}. The comparison results are shown in Table \ref{Table2}. Notably, the proposed method outperforms existing methods in all metrics. Specifically, it achieves 69.29\%, 89.10\%, and 94.06\% accuracy at Rank-1, Rank-5, and Rank-10, respectively, and obtains mAP of 47.15\%. Compared to the second best method APTM \cite{11(57)}, the proposed method shows improvements of 1.07\%, 6.23\%, and 6.56\% at Rank-1, Rank-5, and Rank-10, respectively, and surpasses it by 7.57\% in mAP. These results fully demonstrate the generality and effectiveness of the proposed method.

\textbf{Results on RSTPReid dataset}: We also conduct experiments on RSTPReid dataset and compare the proposed method with 11 state-of-the-art methods, including IMG-Net \cite{43(62)}, AMEN \cite{53(66)}, DSSL \cite{34(25)}, SSAN \cite{3(10)}, SUM \cite{45(35)}, LBUL \cite{48(24)}, IVT \cite{52(22)}, ACSA \cite{61}, CFine \cite{26(38)}, IRRA \cite{10(47)}, and RaSa \cite{28(56)}. The comparison results are shown in Table \ref{Table3}. Observably, the proposed method achieved 48.65\%, 77.15\%, and 87.00\% accuracy at Rank-1, Rank-5, and Rank-10, respectively, and obtained a mAP of 43.08\%. It can also be observed that our method did not surpass CFine \cite{26(38)}, IRRA \cite{10(47)} and RaSa \cite{28(56)} due to the relatively small scale of RSTPReid. Among these three methods, CFine \cite{26(38)} and IRRA\cite{10(47)} utilized pre-trained large-scale models CLIP as backbones, RaSa \cite{28(56)} utilized pre-trained large-scale models ALBEF as backbones, while our proposed method did not use such models. This indicates that when the size of the data set is small, the pre-trained large-scale models can gain a more significant performance improvement. Apart from these specific methods, our proposed method outperformed all other methods. Compared to IVT \cite{52(22)}, it achieved improvements of 1.95\%, 7.15\%, and 8.2\% at Rank-1, Rank-5, and Rank-10, respectively. Nevertheless, the limited size of RSTPReid poses challenges for the proposed method to further improve the retrieval performance.
\begin{table*}[!ht]\small
\centering {\caption{Ablation study on different components of the proposed method on CUHK-PEDES and ICFG-PEDES.}\label{Table4}
\renewcommand\arraystretch{1.4}
\begin{tabular}{c|cc|c|cccc|cccc}
\hline\hline
\multirow{3}*{Methods}  & \multicolumn{3}{c|}{Components}&\multicolumn{4}{c|}{CUHK-PEDES}&\multicolumn{4}{c}{ICFG-PEDES}\\
\cline{2-12}
 & \multicolumn{2}{c|}{PFM}&\multirow{2}*{EKFR}&\multirow{2}*{Rank-1} &\multirow{2}*{Rank-5}&\multirow{2}*{Rank-10}&\multirow{2}*{mAP}&\multirow{2}*{Rank-1} &\multirow{2}*{Rank-5}&\multirow{2}*{Rank-10}&\multirow{2}*{mAP}\\
\cline{2-3}
&Stage \uppercase\expandafter{\romannumeral1}&Stage \uppercase\expandafter{\romannumeral2}&&&&&&&&&\\
\hline
Baseline& & & &51.01&73.85&82.16&48.90&40.56&60.27&69.36&23.93\\
B+PFM$_{1}$ &\checkmark&&&57.68&78.75&85.99&54.58&47.97&67.99&75.90&28.66\\
B+PFM$_{1,2}$ &\checkmark&\checkmark&&59.03&79.27&86.70&55.44&48.48&68.58&76.21&29.50\\
B+EKFR &&&\checkmark&56.14&76.70&84.31&53.10&43.86&63.68&71.79&27.21\\
B+PFM$_{1}$+EKFR &\checkmark&&\checkmark&77.18&94.07&97.35&73.98&66.93&88.38&93.75&45.87\\
\hline
B+PFM+EKFR &\checkmark&\checkmark&\checkmark&77.24&93.71&96.98&73.45&69.29&89.10&94.06&47.15\\
\hline
\end{tabular}}
\end{table*}

\begin{figure}[t!]
\centering
\includegraphics{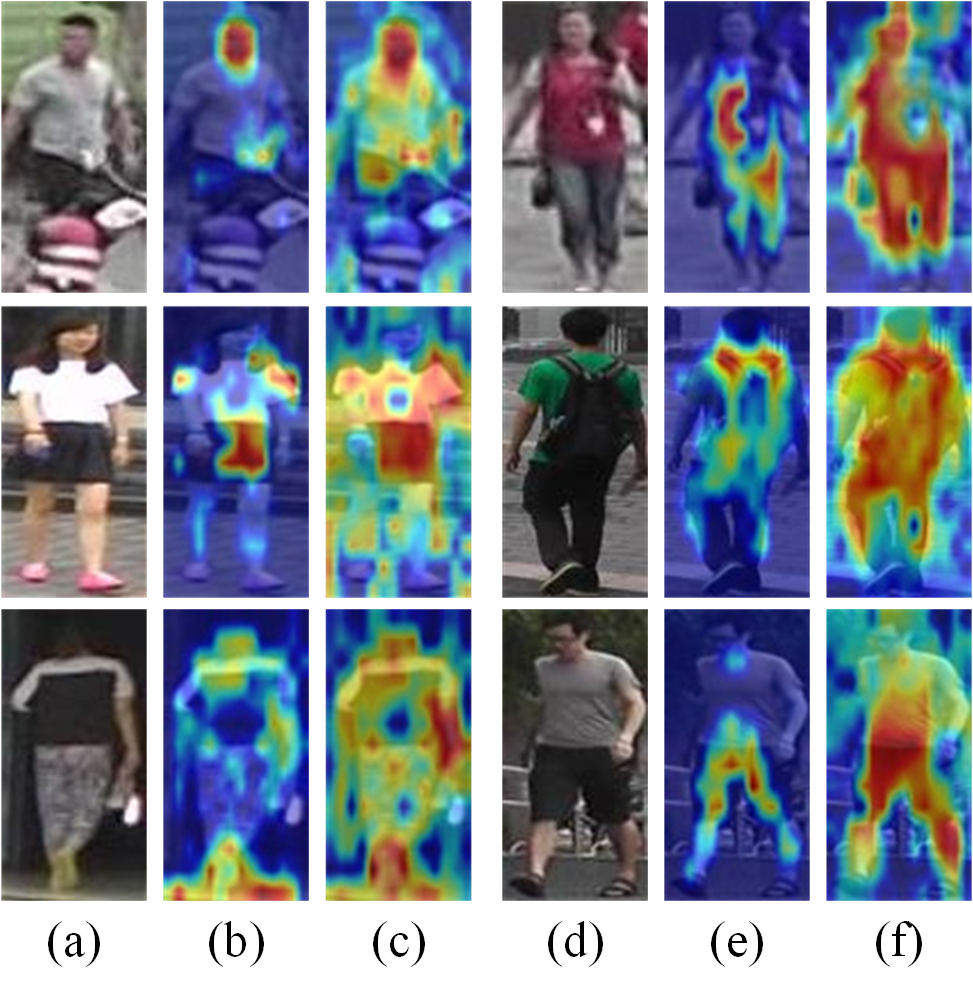}
\caption{The heat map visualization results on CUHK-PEDES dataset are presented as follows: columns (a) and (d) represent the source images, columns (b) and (e)  show the heat maps generated after applying the baseline method,  columns (c) and (f) columns show the heat maps obtained by combining the baseline with the proposed PFM.}
\label{label6}
\end{figure}
\subsection{Ablation Study}
In this section, we conduct extensive ablation experiments to validate the contributions of different components in PFM-EKFP and compare the retrieval performance of different variant models on CUHK-PEDES and ICFG-PEDES datasets. In this process, the model where the ViT and the BERT trained with CMPM and CMPC losses is used as the baseline, and denoted as ``B''. To demonstrate the effectiveness of different components in PFM-EKFP, we add each component sequentially to the baseline to obtain ``B+PFM$_{1}$", ``B+PFM$_{1,2}$'', ``B+EKFR", ``B+PFM$_{1}$+EKFR" and ``B+PFM+EKFR". Specifically, ``B+PFM$_{1}$" means the PFM of stages \uppercase\expandafter{\romannumeral1} added to the baseline, ``B+PFM$_{1,2}$'' means the PFM of stages \uppercase\expandafter{\romannumeral1} and \uppercase\expandafter{\romannumeral2}  added to the baseline, ``B+EKFR" is the baseline plus the EKFR module, ``B+PFM$_{1}$+EKFR" means adding the EKFR module on the basis of ``B+PFM$_{1}$", and ``B+PFM+EKFR" is adding the EKFR module on the basis of ``B+PFM$_{1,2}$'', \emph{i.e.}, the complete model. The experimental results of these variant models on CUHK-PEDES and ICFG-PEDES are reported in Table \ref{Table4}.

\begin{figure}[t!]
\centering
\includegraphics[height=2.2in,width=3.2in]{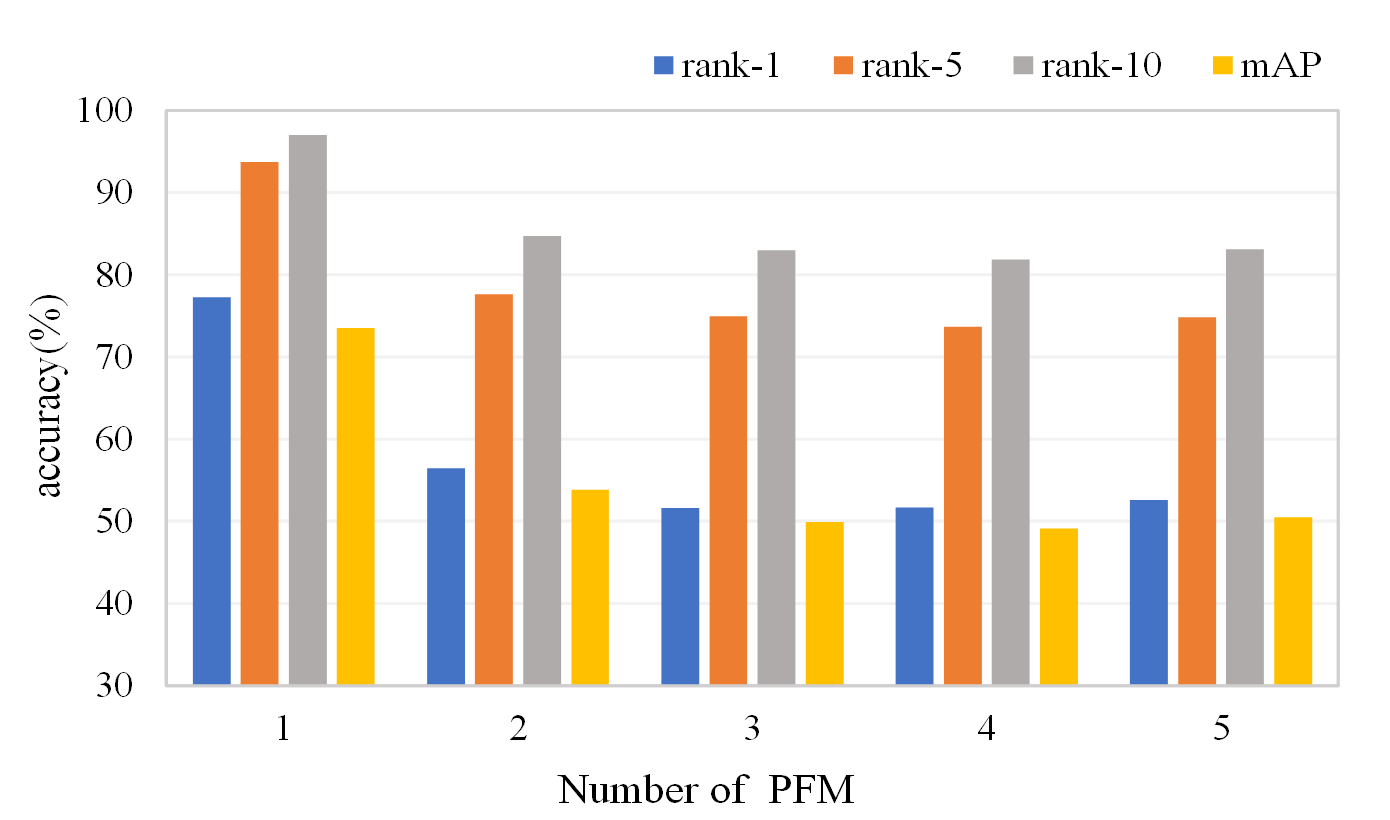}
\caption{The impact of different numbers of PFM on the model performance.}
\label{label7}
\end{figure}

\textbf{Effectiveness of PFM}. From the results in Table \ref{Table4}, the ``B+PFM$_{1,2}$" model improves the Rank-1 performance by 8.02\% and 7.92\% on CUHK-PEDES and ICFG-PEDES datasets compared to the Baseline, respectively. This indicates that the PFM module has a positive impact on the overall model performance. By comparing the results obtained by ``B+PFM$_{1}$'' and ``B+PFM$_{1,2}$'', ``B+PFM$_{1}$+EKFR'' and ``B+PFM+EKFR'', it can be noticed that ``B+PFM$_{1,2}$" outperforms ``B+PFM$_{1}$" by 1.35\% and 0.51\% in terms of Rank-1 accuracy on the two datasets, respectively. Moreover, ``B+PFM+EKFR" improves ``B+PFM$_{1}$+EKFR" by 0.06\% and 2.36\% on the two datasets, respectively. These results suggest that the stage \uppercase\expandafter{\romannumeral2} of PFM plays a positive role in mining discriminative features, and the combination of both stages enables the full potential of the PFM. To further demonstrate the effectiveness of the PFM, we conduct heatmap visualization experiment on CUHK-PEDES dataset. As shown in Fig. \ref{label6}, it can be observed that, with only the baseline, the network focuses on less discriminative information. While after adding the PFM, the network pay attention to more discriminative information, indicating the effectiveness of the PFM for extracting discriminative features.
 \begin{figure}[!t] \centering
 \subfigure[]  {\includegraphics[height=1.3in,width=1.7in,angle=0]{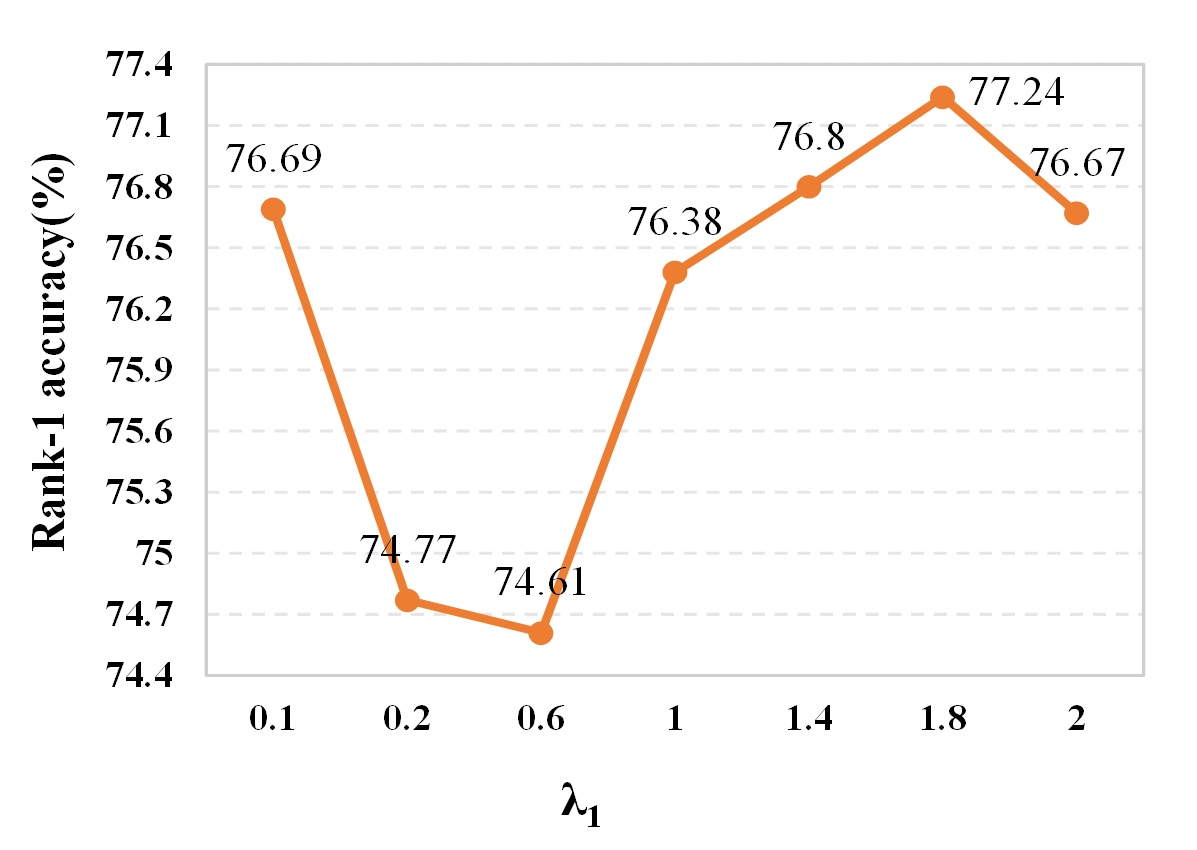}}
\subfigure[]   {\includegraphics[height=1.3in,width=1.7in,angle=0]{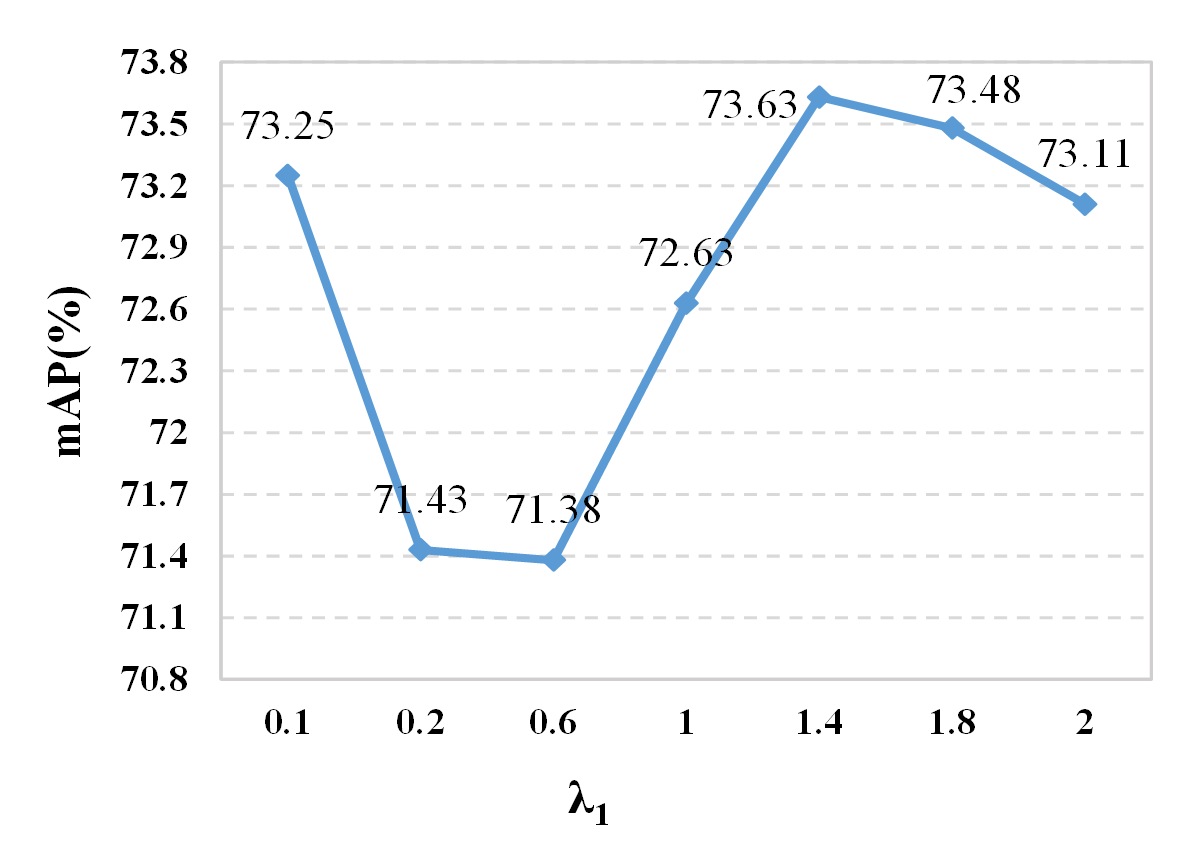}}
  \caption{Effect of $\lambda_{1}$ on model performance at different values. (a) is the effect of $\lambda_{1}$ on Rank-1 accuracy at different values; (b) is the effect of $\lambda_{1}$ on mAP at different values.}
 \label{label8}
 \end{figure}
 \begin{figure}[!t] \centering
 \subfigure[]  {\includegraphics[height=1.3in,width=1.7in,angle=0]{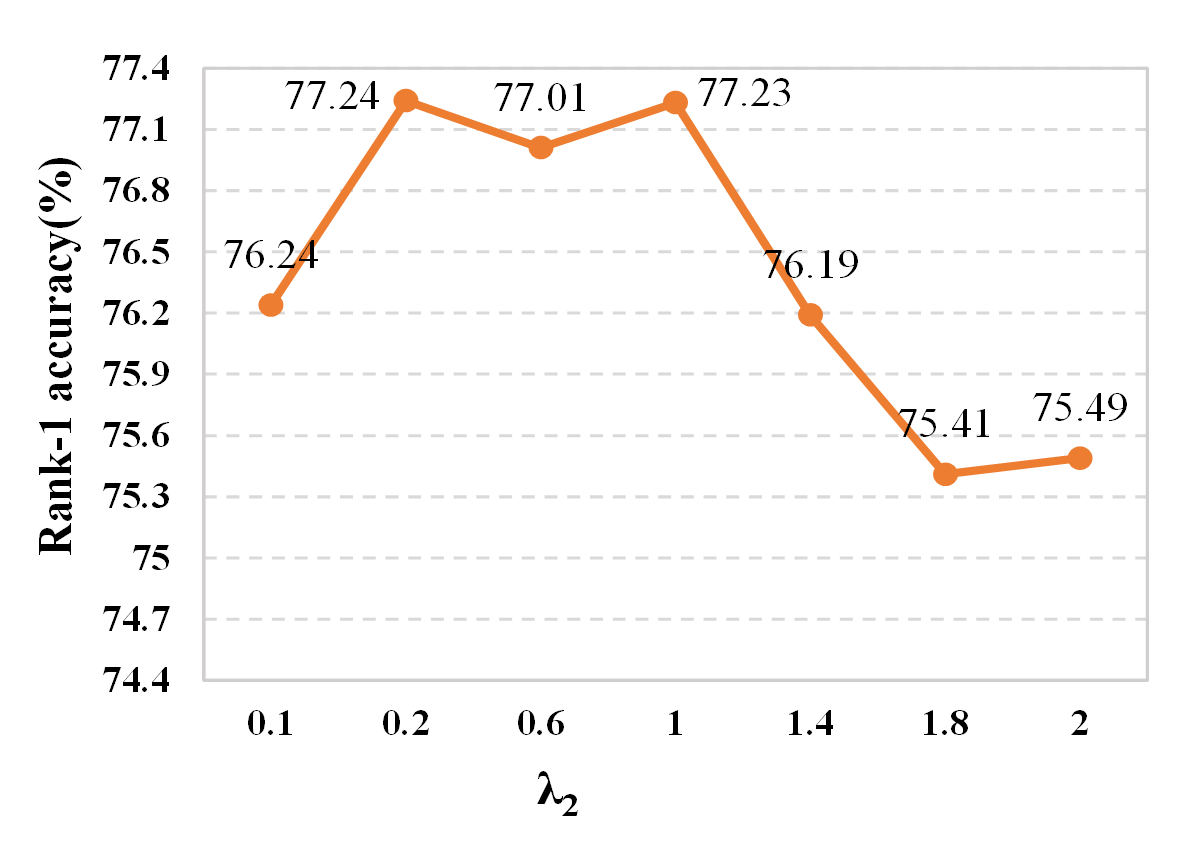}}
\subfigure[]   {\includegraphics[height=1.3in,width=1.7in,angle=0]{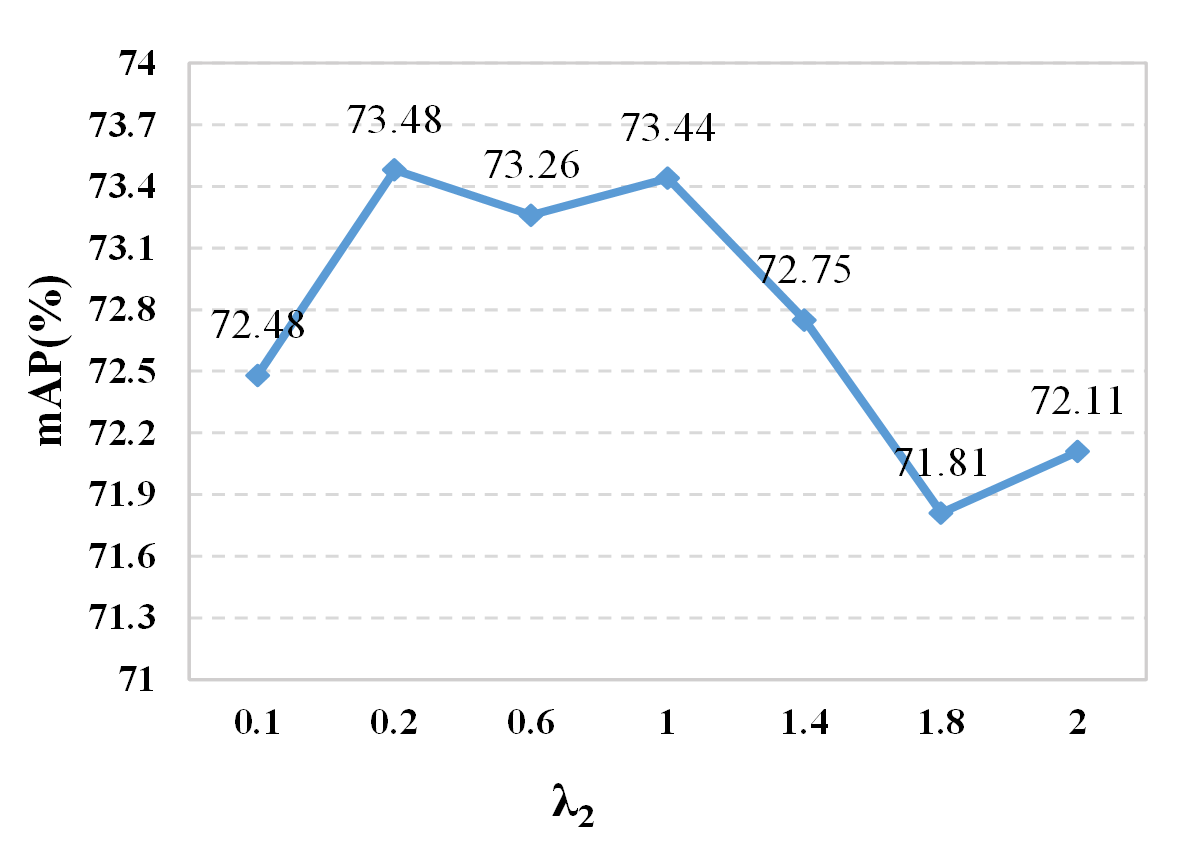}}
  \caption{Effect of $\lambda_{2}$ on model performance at different values. (a) is the effect of $\lambda_{2}$ on Rank-1 accuracy at different values; (b) is the effect of $\lambda_{2}$ on mAP at different values.}
  \label{label9}
 \end{figure}

\begin{figure*}[t!]
\centering
\includegraphics{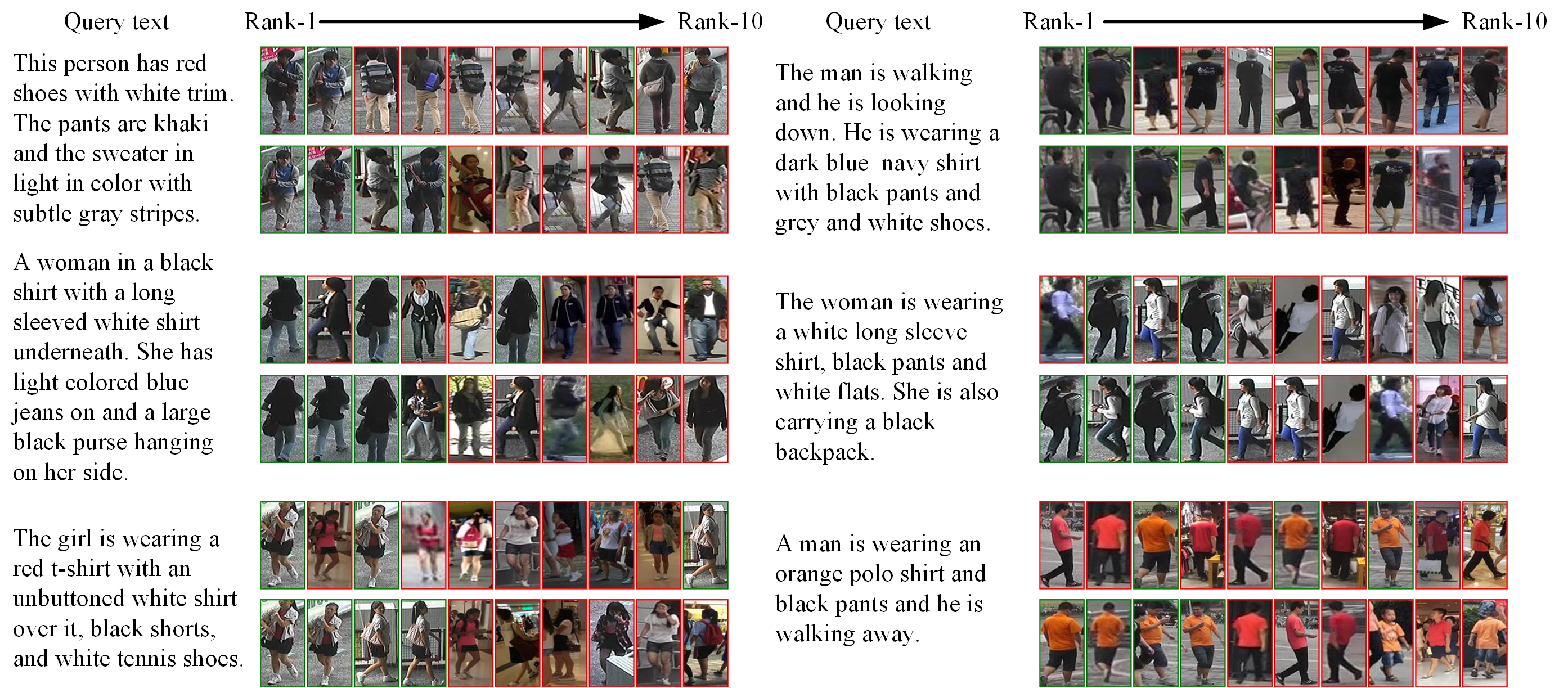}
\caption{Six retrieval examples on CUHK-PEDES dataset are displayed. The first row in each example represents the top-10 retrieval results obtained by the baseline, and the second row represents the top-10 retrieval results obtained by the method proposed. The green bounding box on the image indicates the retrieval result is correct, while the red bounding box indicates the retrieval result is incorrect.}
\label{label10}
\end{figure*}

\textbf{Effectiveness of EKFR}. Based on the experimental results in Table \ref{Table4}, on Rank-1, the ``B+EKFR" outperforms the Baseline by 5.13\% and 3.30\%  on the two datasets, respectively. The ``B+PFM$_{1}$+EKFR" shows improvements of 19.50\% and 18.96\% over ``B+PFM$_{1}$" on the two datasets, respectively. Moreover, the ``B+PFM+EKFR" significantly improves the matching accuracy by 18.21\% and 20.81\% over ``B+PFM$_{1,2}$" on the two datasets, respectively. These results indicate that EKFR contributes to improving the matching performance, and the best result is achieved by executing EKFR on the basis of PFM. It shows that the extraction of image and text discriminative features is extremely important for subsequent further feature optimization.

\textbf{Effectiveness of PFM+EKFR}. By comparing the experimental results in Table \ref{Table4}, we can observe that the matching accuracy on Rank-1 achieved by ``B+PFM+EKFR" outperforms that achieved by Baseline by 26.23\% and 28.73\% on the two datasets, respectively. This indicates that the proposed complete model achieves the best experimental performance.

In summary, the experimental results in Table \ref{Table4} demonstrate the effectiveness of the PFM and EKFR methods in the entire network. Their incorporation significantly improves the retrieval performance, and the effect is the best when they are used in combination.

\subsection{Further Discussion}
This study proposes a progressive feature mining method and its effectiveness for discriminative feature mining can be seen in Table \ref{Table4}. To study the impact of multiple PFM modules on the entire network, we stack  1, 2, 3, 4, and 5 PFM modules in the model separately, and then conduct experiments on CUHK-PEDES dataset. The experimental results are shown in Fig. \ref{label7}. It can be seen that the performance is the best when the number of PFM modules is 1, and the performance of the model decreases after adding PFM modules. This indicates that one PFM is enough.
\subsection{Parameter Selection and Analysis}
In this section, we analyze the influence of the hyperparameters $\lambda_{1}$ and $\lambda_{2}$ in Eq.(22) on the retrieval performance. When analyzing the influence of one of the hyperparameters, the other one is fixed.

\textbf{The influence of $\lambda_{1}$}.
Fig. \ref{label8} shows the impact of $\lambda_{1}$ on the model performance on CUHK-PEDES dataset, where Fig. \ref{label8}(a) and Fig. \ref{label8}(b) show the impact of $\lambda_{1}$ on Rank-1 and mAP, respectively. It can be seen that when $\lambda_{1}$ = 0.6, the Rank-1 and mAP are the lowest; when $\lambda_{1}$ = 1.4, the mAP reaches a peak value of 73.63\%; and when $\lambda_{1}$ = 1.8, Rank-1 reaches a peak value of 77.24\%. These indicate that with the value of $\lambda_{1}$ selected within $\left[1.4,1.8\right]$, the model performance will be better. Therefore, we set $\lambda_{1}$ to 1.8 throughout the experiments.

\textbf{The influence of $\lambda_{2}$}.
We fixed $\lambda_{1}$=1.8 to analyze the impact of $\lambda_{2}$ on the model performance. As shown in Fig. \ref{label9}(a) and Fig. \ref{label9}(b), when $\lambda_{2}$ = 0.2,  Rank-1 and mAP reach their respective highest values, namely, Rank-1 reaches 77.24\% and mAP reaches 73.48\%. When $\lambda_{2}$ = 1, the Rank-1 and mAP are close to that of $\lambda_{2}$ = 0.2, while when $\lambda_{2}$ varies from 1 to 2, the performance gradually decreases, and the Rank-1 and mAP decline to 75.49\% and 72.11\%, respectively. Therefore, throughout the experiment, $\lambda_{2}$ is set to 0.2.

\subsection{ Qualitative Results}
Fig. \ref{label10} shows the retrieval results of different models on CUHK-PEDES dataset. It is evident that the proposed method obtains correct matches in rank-4, while the rank-4 results obtained by the baseline only contain 1 or 2 correct matching results. This demonstrates that the proposed method is effective in the text-pedestrian image retrieval task. From the retrieval results, it is also noticeable that the rank-10 results contain incorrectly retrieved images due to the high similarity between the incorrectly retrieved image and the correctly retrieved image. In terms of appearance, these pedestrians share similar attributes, and the difference only lies in fine-grained aspects, such as the difference between long pants and shorts. Enhancing the discrimination power of fine-grained features will facilitate the differentiation of hard negative samples. This will be what we will explore in the future.

\section{Conclusion}
In this paper, we propose a progressive feature mining and external knowledge-assisted feature purification method for text-pedestrian image retrieval, which includes PFM and EKFR sub-methods. PFM mines the discriminative features of image and text in a progressive manner to improve the expressive ability of features; EKFR uses external knowledge to enhance identity-consistent features and weaken identity-inconsistent features, thereby realizing the purification of features, alleviating the disturbance of text diversity and same-modal negative sample correlation features on cross-modal matching. The model structure and loss functions are discussed in detail. The experimental results on the three data sets demonstrate the effectiveness of the proposed method. The ablation study verifies the effectiveness of each sub-module. In future work, we will focus on mining fine-grained discriminative features to further improve the representation of features, thereby improving retrieval performance.

\bibliography{mybibfile1}

\begin{thebibliography}{10}
\providecommand{\url}[1]{#1}
\csname url@samestyle\endcsname
\providecommand{\newblock}{\relax}
\providecommand{\bibinfo}[2]{#2}
\providecommand{\BIBentrySTDinterwordspacing}{\spaceskip=0pt\relax}
\providecommand{\BIBentryALTinterwordstretchfactor}{4}
\providecommand{\BIBentryALTinterwordspacing}{\spaceskip=\fontdimen2\font plus
\BIBentryALTinterwordstretchfactor\fontdimen3\font minus
  \fontdimen4\font\relax}
\providecommand{\BIBforeignlanguage}[2]{{%
\expandafter\ifx\csname l@#1\endcsname\relax
\typeout{** WARNING: IEEEtran.bst: No hyphenation pattern has been}%
\typeout{** loaded for the language `#1'. Using the pattern for}%
\typeout{** the default language instead.}%
\else
\language=\csname l@#1\endcsname
\fi
#2}}
\providecommand{\BIBdecl}{\relax}
\BIBdecl

\bibitem{56}
H.~Li, Y.~Chen, D.~Tao, Z.~Yu, and G.~Qi, ``Attribute-aligned domain-invariant
  feature learning for unsupervised domain adaptation person
  re-identification,'' \emph{IEEE Transactions on Information Forensics and
  Security}, vol.~16, pp. 1480--1494, 2020.

\bibitem{57}
Y.~Zhang, Y.~Wang, H.~Li, and S.~Li, ``Cross-compatible embedding and semantic
  consistent feature construction for sketch re-identification,'' in
  \emph{Proceedings of the 30th ACM International Conference on Multimedia},
  2022, pp. 3347--3355.

\bibitem{63}
Y.~Feng, Y.~Ji, F.~Wu, G.~Gao, Y.~Gao, T.~Liu, S.~Liu, X.-Y. Jing, and J.~Luo,
  ``Occluded visible-infrared person re-identification,'' \emph{IEEE
  Transactions on Multimedia}, 2022.

\bibitem{58}
H.~Li, S.~Yan, Z.~Yu, and D.~Tao, ``Attribute-identity embedding and
  self-supervised learning for scalable person re-identification,'' \emph{IEEE
  Transactions on Circuits and Systems for Video Technology}, vol.~30, no.~10,
  pp. 3472--3485, 2019.

\bibitem{65}
Z.~Zhang, C.~Lan, W.~Zeng, Z.~Chen, and S.-F. Chang, ``Beyond triplet loss:
  Meta prototypical n-tuple loss for person re-identification,'' \emph{IEEE
  Transactions on Multimedia}, vol.~24, pp. 4158--4169, 2021.

\bibitem{59}
S.~Li, F.~Li, J.~Li, H.~Li, B.~Zhang, D.~Tao, and X.~Gao, ``Logical relation
  inference and multiview information interaction for domain adaptation person
  re-identification,'' \emph{IEEE Transactions on Neural Networks and Learning
  Systems}, 2023.

\bibitem{64}
Z.~Yu, Y.~Zhao, B.~Hong, Z.~Jin, J.~Huang, D.~Cai, and X.-S. Hua,
  ``Apparel-invariant feature learning for person re-identification,''
  \emph{IEEE Transactions on Multimedia}, vol.~24, pp. 4482--4492, 2021.

\bibitem{60}
H.~Li, N.~Dong, Z.~Yu, D.~Tao, and G.~Qi, ``Triple adversarial learning and
  multi-view imaginative reasoning for unsupervised domain adaptation person
  re-identification,'' \emph{IEEE Transactions on Circuits and Systems for
  Video Technology}, vol.~32, no.~5, pp. 2814--2830, 2021.

\bibitem{1(7)}
K.~Niu, Y.~Huang, W.~Ouyang, and L.~Wang, ``Improving description-based person
  re-identification by multi-granularity image-text alignments,'' \emph{IEEE
  Transactions on Image Processing}, vol.~29, pp. 5542--5556, 2020.

\bibitem{2(8)}
C.~Gao, G.~Cai, X.~Jiang, F.~Zheng, J.~Zhang, Y.~Gong, P.~Peng, X.~Guo, and
  X.~Sun, ``Contextual non-local alignment over full-scale representation for
  text-based person search,'' \emph{arXiv preprint arXiv:2101.03036}, 2021.

\bibitem{3(10)}
Z.~Ding, C.~Ding, Z.~Shao, and D.~Tao, ``Semantically self-aligned network for
  text-to-image part-aware person re-identification,'' 2021, arXiv preprint
  arXiv:2101.03036.

\bibitem{4(12)}
Y.~Chen, G.~Zhang, Y.~Lu, Z.~Wang, and Y.~Zheng, ``Tipcb: A simple but
  effective part-based convolutional baseline for text-based person search,''
  \emph{Neurocomputing}, vol. 494, pp. 171--181, 2022.

\bibitem{5(18)}
Y.~Jing, W.~Wang, L.~Wang, and T.~Tan, ``Learning aligned image-text
  representations using graph attentive relational network,'' \emph{IEEE
  Transactions on Image Processing}, vol.~30, pp. 1840--1852, 2021.

\bibitem{6(4)}
S.~Li, M.~Cao, and M.~Zhang, ``Learning semantic-aligned feature representation
  for text-based person search,'' in \emph{ICASSP 2022 - 2022 IEEE
  International Conference on Acoustics, Speech and Signal Processing
  (ICASSP)}, 2022, pp. 2724--2728.

\bibitem{7(11)}
Z.~Shao, X.~Zhang, M.~Fang, Z.~Lin, J.~Wang, and C.~Ding, ``Learning
  granularity-unified representations for text-to-image person
  re-identification,'' in \emph{Proceedings of the 30th ACM International
  Conference on Multimedia}, 2022, pp. 5566--5574.

\bibitem{8(37)}
S.~Yan, H.~Tang, L.~Zhang, and J.~Tang, ``Image-specific information
  suppression and implicit local alignment for text-based person search,''
  \emph{arXiv preprint arXiv:2208.14365}, 2022.

\bibitem{9(42)}
F.~Li, H.~Zhou, H.~Li, Y.~Zhang, and Z.~Yu, ``Person text-image matching via
  text-featur interpretability embedding and external attack node
  implantation,'' \emph{arXiv preprint arXiv:2211.08657}, 2022.

\bibitem{10(47)}
D.~Jiang and M.~Ye, ``Cross-modal implicit relation reasoning and aligning for
  text-to-image person retrieval,'' in \emph{Proceedings of the IEEE/CVF
  Conference on Computer Vision and Pattern Recognition}, 2023, pp. 2787--2797.

\bibitem{11(57)}
S.~Yang, Y.~Zhou, Y.~Wang, Y.~Wu, L.~Zhu, and Z.~Zheng, ``Towards unified
  text-based person retrieval: A large-scale multi-attribute and language
  search benchmark,'' \emph{arXiv preprint arXiv:2306.02898}, 2023.

\bibitem{12(14)}
Y.~Jing, C.~Si, J.~Wang, W.~Wang, L.~Wang, and T.~Tan, ``Pose-guided joint
  global and attentive local matching network for text-based person search,''
  \emph{Association for the Advance of Artificial Intelligence (AAAI)}, 2020.

\bibitem{13(19)}
Z.~Wang, Z.~Fang, J.~Wang, and Y.~Yang, ``Vitaa: Visual-textual attributes
  alignment in person search by natural language,'' in \emph{Computer Vision --
  ECCV 2020}, 2020, pp. 402--420.

\bibitem{14(29)}
Y.~Jing, C.~Si, J.~Wang, W.~Wang, L.~Wang, and T.~Tan, ``Pose-guided
  multi-granularity attention network for text-based person search,'' in
  \emph{Proceedings of the AAAI Conference on Artificial Intelligence},
  vol.~34, no.~07, 2020, pp. 11\,189--11\,196.

\bibitem{15(27)}
S.~Aggarwal, R.~V. Babu, and A.~Chakraborty, ``Text-based person search via
  attribute-aided matching,'' in \emph{2020 IEEE Winter Conference on
  Applications of Computer Vision (WACV)}, 2020, pp. 2606--2614.

\bibitem{16(32)}
C.~Wang, Z.~Luo, Y.~Lin, and S.~Li, ``Improving embedding learning by virtual
  attribute decoupling for text-based person search,'' \emph{Neural Computing
  and Applications}, pp. 1--23, 2022.

\bibitem{55}
H.~Tang, C.~Yuan, Z.~Li, and J.~Tang, ``Learning attention-guided pyramidal
  features for few-shot fine-grained recognition,'' \emph{Pattern Recognition},
  vol. 130, p. 108792, 2022.

\bibitem{18(2-1)}
Z.~Zheng, L.~Zheng, M.~Garrett, Y.~Yang, M.~Xu, and Y.~Shen, ``Dual-path
  convolutional image-text embeddings with instance loss,'' \emph{ACM
  Transactions on Multimedia Computing, Communications, and Applications
  (TOMM)}, 2020.

\bibitem{19(2-2)}
N.~Messina, F.~Falchi, A.~Esuli, and G.~Amato, ``Transformer reasoning network
  for image- text matching and retrieval,'' in \emph{2020 25th International
  Conference on Pattern Recognition (ICPR)}, 2021, pp. 5222--5229.

\bibitem{20(2-8)}
Z.~Ji, K.~Chen, and H.~Wang, ``Step-wise hierarchical alignment network for
  image-text matching,'' \emph{arXiv preprint arXiv:2106.06509}, 2021.

\bibitem{21(2-10)}
Z.~Ji, H.~Wang, J.~Han, and Y.~Pang, ``Sman: Stacked multimodal attention
  network for cross-modal image–text retrieval,'' \emph{IEEE Transactions on
  Cybernetics}, vol.~52, no.~2, pp. 1086--1097, 2022.

\bibitem{22(2-14)}
K.~Li, Y.~Zhang, K.~Li, Y.~Li, and Y.~Fu, ``Visual semantic reasoning for
  image-text matching,'' in \emph{2019 IEEE/CVF International Conference on
  Computer Vision (ICCV)}, 2019, pp. 4653--4661.

\bibitem{23(2-13)}
X.~Liu, Y.~He, Y.-M. Cheung, X.~Xu, and N.~Wang, ``Learning
  relationship-enhanced semantic graph for fine-grained image-text matching,''
  \emph{IEEE Transactions on Cybernetics}, pp. 1--14, 2022.

\bibitem{24(2-16)}
X.~Ge, F.~Chen, S.~Xu, F.~Tao, and J.~M. Jose, ``Cross-modal semantic enhanced
  interaction for image-sentence retrieval,'' in \emph{2023 IEEE/CVF Winter
  Conference on Applications of Computer Vision (WACV)}, 2023, pp. 1022--1031.

\bibitem{17(0)}
S.~Li, T.~Xiao, H.~Li, B.~Zhou, D.~Yue, and X.~Wang, ``Person search with
  natural language description,'' in \emph{2017 IEEE Conference on Computer
  Vision and Pattern Recognition (CVPR)}, 2017, pp. 5187--5196.

\bibitem{25(65)}
S.~Bird, ``Nltk: the natural language toolkit,'' in \emph{Proceedings of the
  COLING/ACL 2006 Interactive Presentation Sessions}, 2006, pp. 69--72.

\bibitem{26(38)}
S.~Yan, N.~Dong, L.~Zhang, and J.~Tang, ``Clip-driven fine-grained text-image
  person re-identification,'' \emph{arXiv preprint arXiv:2210.10276}, 2022.

\bibitem{27(2-5)}
A.~Radford, J.~W. Kim, C.~Hallacy, A.~Ramesh, G.~Goh, S.~Agarwal, G.~Sastry,
  A.~Askell, P.~Mishkin, J.~Clark \emph{et~al.}, ``Learning transferable visual
  models from natural language supervision,'' in \emph{International conference
  on machine learning}.\hskip 1em plus 0.5em minus 0.4em\relax PMLR, 2021, pp.
  8748--8763.

\bibitem{28(56)}
Y.~Bai, M.~Cao, D.~Gao, Z.~Cao, C.~Chen, Z.~Fan, L.~Nie, and M.~Zhang, ``Rasa:
  Relation and sensitivity aware representation learning for text-based person
  search,'' \emph{arXiv preprint arXiv:2305.13653}, 2023.

\bibitem{29(2-19)}
J.~Li, R.~Selvaraju, A.~Gotmare, S.~Joty, C.~Xiong, and S.~C.~H. Hoi, ``Align
  before fuse: Vision and language representation learning with momentum
  distillation,'' \emph{Advances in neural information processing systems},
  vol.~34, pp. 9694--9705, 2021.

\bibitem{31(63)}
A.~Dosovitskiy, L.~Beyer, A.~Kolesnikov, D.~Weissenborn, X.~Zhai,
  T.~Unterthiner, M.~Dehghani, M.~Minderer, G.~Heigold, S.~Gelly \emph{et~al.},
  ``An image is worth 16x16 words: Transformers for image recognition at
  scale,'' \emph{arXiv preprint arXiv:2010.11929}, 2020.

\bibitem{30(59)}
O.~Russakovsky, J.~Deng, H.~Su, J.~Krause, S.~Satheesh, S.~Ma, Z.~Huang,
  A.~Karpathy, A.~Khosla, M.~Bernstein \emph{et~al.}, ``Imagenet large scale
  visual recognition challenge,'' \emph{International journal of computer
  vision}, vol. 115, pp. 211--252, 2015.

\bibitem{32(64)}
J.~Devlin, M.-W. Chang, K.~Lee, and K.~Toutanova, ``Bert: Pre-training of deep
  bidirectional transformers for language understanding,'' \emph{arXiv preprint
  arXiv:1810.04805}, 2018.

\bibitem{33(49)}
Y.~Zhang and H.~Lu, ``Deep cross-modal projection learning for image-text
  matching,'' in \emph{Proceedings of the European conference on computer
  vision (ECCV)}, 2018, pp. 686--701.

\bibitem{54}
A.~Vaswani, N.~Shazeer, N.~Parmar, J.~Uszkoreit, L.~Jones, A.~N. Gomez,
  {\L}.~Kaiser, and I.~Polosukhin, ``Attention is all you need,''
  \emph{Advances in neural information processing systems}, vol.~30, 2017.

\bibitem{34(25)}
A.~Zhu, Z.~Wang, Y.~Li, X.~Wan, J.~Jin, T.~Wang, F.~Hu, and G.~Hua, ``Dssl:
  Deep surroundings-person separation learning for text-based person
  retrieval,'' in \emph{Proceedings of the 29th ACM International Conference on
  Multimedia}, 2021, pp. 209--217.

\bibitem{39(50)}
Y.~Wang, C.~Bo, D.~Wang, S.~Wang, Y.~Qi, and H.~Lu, ``Language person search
  with mutually connected classification loss,'' in \emph{ICASSP 2019-2019 IEEE
  International Conference on Acoustics, Speech and Signal Processing
  (ICASSP)}.\hskip 1em plus 0.5em minus 0.4em\relax IEEE, 2019, pp. 2057--2061.

\bibitem{40(26)}
J.~Liu, Z.-J. Zha, R.~Hong, M.~Wang, and Y.~Zhang, ``Deep adversarial graph
  attention convolution network for text-based person search,'' in
  \emph{Proceedings of the 27th ACM International Conference on Multimedia},
  2019, pp. 665--673.

\bibitem{41(9)}
N.~Sarafianos, X.~Xu, and I.~A. Kakadiaris, ``Adversarial representation
  learning for text-to-image matching,'' in \emph{Proceedings of the IEEE/CVF
  international conference on computer vision}, 2019, pp. 5814--5824.

\bibitem{38(48)}
Z.~Zheng, L.~Zheng, M.~Garrett, Y.~Yang, M.~Xu, and Y.-D. Shen, ``Dual-path
  convolutional image-text embeddings with instance loss,'' \emph{ACM
  Transactions on Multimedia Computing, Communications, and Applications
  (TOMM)}, vol.~16, no.~2, pp. 1--23, 2020.

\bibitem{62}
Z.-J. Zha, J.~Liu, D.~Chen, and F.~Wu, ``Adversarial attribute-text embedding
  for person search with natural language query,'' \emph{IEEE Transactions on
  Multimedia}, vol.~22, no.~7, pp. 1836--1846, 2020.

\bibitem{43(62)}
Z.~Wang, A.~Zhu, Z.~Zheng, J.~Jin, Z.~Xue, and G.~Hua, ``Img-net:
  inner-cross-modal attentional multigranular network for description-based
  person re-identification,'' \emph{Journal of Electronic Imaging}, vol.~29,
  no.~4, pp. 043\,028--043\,028, 2020.

\bibitem{44(30)}
K.~Zheng, W.~Liu, J.~Liu, Z.-J. Zha, and T.~Mei, ``Hierarchical gumbel
  attention network for text-based person search,'' in \emph{Proceedings of the
  28th ACM International Conference on Multimedia}, 2020, pp. 3441--3449.

\bibitem{42(51)}
Y.~Chen, R.~Huang, H.~Chang, C.~Tan, T.~Xue, and B.~Ma, ``Cross-modal knowledge
  adaptation for language-based person search,'' \emph{IEEE Transactions on
  Image Processing}, vol.~30, pp. 4057--4069, 2021.

\bibitem{46(1)}
C.~Wang, Z.~Luo, Y.~Lin, and S.~Li, ``Text-based person search via
  multi-granularity embedding learning.'' in \emph{IJCAI}, 2021, pp.
  1068--1074.

\bibitem{47(20)}
Y.~Wu, Z.~Yan, X.~Han, G.~Li, C.~Zou, and S.~Cui, ``Lapscore: language-guided
  person search via color reasoning,'' in \emph{Proceedings of the IEEE/CVF
  International Conference on Computer Vision}, 2021, pp. 1624--1633.

\bibitem{45(35)}
Z.~Wang, A.~Zhu, J.~Xue, D.~Jiang, C.~Liu, Y.~Li, and F.~Hu, ``Sum: Serialized
  updating and matching for text-based person retrieval,''
  \emph{Knowledge-Based Systems}, vol. 248, p. 108891, 2022.

\bibitem{61}
Z.~Ji, J.~Hu, D.~Liu, L.~Y. Wu, and Y.~Zhao, ``Asymmetric cross-scale alignment
  for text-based person search,'' \emph{IEEE Transactions on Multimedia}, 2022.

\bibitem{48(24)}
Z.~Wang, A.~Zhu, J.~Xue, X.~Wan, C.~Liu, T.~Wang, and Y.~Li, ``Look before you
  leap: Improving text-based person retrieval by learning a consistent
  cross-modal common manifold,'' in \emph{Proceedings of the 30th ACM
  International Conference on Multimedia}, 2022, pp. 1984--1992.

\bibitem{50(23)}
Z.~Wang, A.~Zhu, J.~Xue, X.~Wan, C.~Liu, and T.~Wang, ``Caibc: Capturing
  all-round information beyond color for text-based person retrieval,'' in
  \emph{Proceedings of the 30th ACM International Conference on Multimedia},
  2022, pp. 5314--5322.

\bibitem{51(21)}
A.~Farooq, M.~Awais, J.~Kittler, and S.~S. Khalid, ``Axm-net: Implicit
  cross-modal feature alignment for person re-identification,'' in
  \emph{Proceedings of the AAAI Conference on Artificial Intelligence},
  vol.~36, no.~4, 2022, pp. 4477--4485.

\bibitem{52(22)}
X.~Shu, W.~Wen, H.~Wu, K.~Chen, Y.~Song, R.~Qiao, B.~Ren, and X.~Wang, ``See
  finer, see more: Implicit modality alignment for text-based person
  retrieval,'' in \emph{European Conference on Computer Vision}.\hskip 1em plus
  0.5em minus 0.4em\relax Springer, 2022, pp. 624--641.

\bibitem{49(13)}
X.~Han, S.~He, L.~Zhang, and T.~Xiang, ``Text-based person search with limited
  data,'' \emph{arXiv preprint arXiv:2110.10807}, 2021.

\bibitem{35(60)}
D.~P. Kingma and J.~Ba, ``Adam: A method for stochastic optimization,''
  \emph{arXiv preprint arXiv:1412.6980}, 2014.

\bibitem{36(61)}
H.~Luo, Y.~Gu, X.~Liao, S.~Lai, and W.~Jiang, ``Bag of tricks and a strong
  baseline for deep person re-identification,'' in \emph{Proceedings of the
  IEEE/CVF conference on computer vision and pattern recognition workshops},
  2019, pp. 0--0.

\bibitem{53(66)}
Z.~Wang, J.~Xue, A.~Zhu, Y.~Li, M.~Zhang, and C.~Zhong, ``Amen: Adversarial
  multi-space embedding network for text-based person re-identification,'' in
  \emph{Pattern Recognition and Computer Vision: 4th Chinese Conference, PRCV
  2021, Beijing, China, October 29--November 1, 2021, Proceedings, Part II
  4}.\hskip 1em plus 0.5em minus 0.4em\relax Springer, 2021, pp. 462--473.

\end{thebibliography}
\end{document}